\documentclass[review]{elsarticle}
\usepackage{framed,multirow}
\usepackage{amsmath}
\usepackage{amssymb}
\usepackage{latexsym}
\usepackage{hyperref} 
\usepackage{url}
\usepackage{xcolor}
\definecolor{newcolor}{rgb}{.8,.349,.1}
\usepackage{graphicx}
\graphicspath{{./Images/}}
\DeclareGraphicsExtensions{.pdf,.jpg,.png,.eps}
\usepackage{float}
\usepackage{multicol, blindtext}
\usepackage{color,soul}
\usepackage{upgreek}
\usepackage{tablefootnote}
\usepackage{threeparttable}
\usepackage{array}
\usepackage{multirow}
\newenvironment{rcases}
{\left.\begin{aligned}}
	{\end{aligned}\right\rbrace}
\usepackage[utf8]{inputenc}
\usepackage[caption=false]{subfig}
\usepackage{lineno,hyperref}
\modulolinenumbers[5]
\usepackage{lipsum}
\usepackage[numbers]{natbib}
\usepackage{algorithm}
\usepackage{algpseudocode}
\usepackage{algorithmicx}
\usepackage[utf8]{inputenc}
\algdef{SE}[DOWHILE]{Do}{doWhile}{\algorithmicdo}[1]{\algorithmicwhile\ #1}%
\journal{Engineering Application of Artificial Intillegence}

\pdfoutput=1
\begin{document}

\begin{frontmatter}

\title{Dual approach for object tracking based on optical flow and swarm intelligence}

\author{Mr.Rajesh Misra}
\address{S.A.Jaipuria College,10 Raja Naba Krishna Street, Shobhabazar, Kolkata,India}
\ead{rajeshmisra.85@gmail.com}

\author{Dr.Kumar S Ray}
\address{Indian Statistical Institute, 203 B.T.Road, Kolkata-108, India}
\ead{ksray@isical.ac.in}



\begin{abstract}
Though object tracking is a very old problem still there are several challenges to be solved;for instance, variation of illumination of light, noise, occlusion, sudden start and stop of moving object, shading etc.In this paper we propose a dual approach for object tracking based on optical flow and swarm Intelligence.The optical flow based KLT tracker, tracks the dominant points of the target object from first frame to last frame of a video sequence;whereas swarm Intelligence based PSO tracker simultaneously tracks the boundary information of the target object from second frame to last frame of the same video sequence.The boundary information of the target object is captured by the polygonal approximation of the same.The dual approach to object tracking is inherently robust with respect to the above stated problems.We compare the performance of the proposed dual tracking algorithm with several benchmark datasets and in most of the cases we obtain superior results.
\end{abstract}


\end{frontmatter}

%\linenumbers

\section{Introduction}

Object Tracking employs the idea of following an object as long as its movement can be captured by a camera in various environments under Variable Background and Static Background. Moving object detection and tracking pose a challenge in real world scenarios like automatic surveillance system, traffic monitoring, vehicle navigation etc. In many scenarios where background changes dynamically due to motion of camera, abrupt changes in speed of the tracked object, change in illumination of light, noise, occlusion etc., tracking becomes very complex and challenging. Therefore tracking algorithm under such situation should be robust, flexible and adaptive. It should be capable of real time execution.

Moving object tracking is existing for past several decades. Many methods have been proposed with a certain degree of accuracy and effectiveness. Still there remains several challenging problems in tracking due to the reasons stated earlier. In this paper we adopt a dual tracking approach based on optical flow and swarm intelligence so that the tracking becomes very robust under the challenges as stated.

Optical flow is the pattern of motion of images between two consecutive frames generated by movement of object or camera. The resultant vector of the optical flow is the displacement vector containing position of pixels from first frame to second frame. This optical flow provides a good amount of motion information of moving object, and thus encourage researchers to apply that information in moving object detection as well as tracking. There exist several optical flow methods, like Lucas-–Kanade method \citep{Tom91}, Horn-–Schunck method \citep{Hor93}, Buxton–-Buxton method\citep{Bux84}, Black–-Jepson method\citep{Jep93} etc. Among all these methods on optical flow \cite{Bar94}, Horn-Schunck and Lucas-–Kanade are more popular than others. Both these methods have their own merits and demerits. Shin et.al \citep{shin05} proposed optical flow based object tracking under non-prior training active feature model by localizing of an object-of-interest then applying correction based model using spatio-temporal information and then applying NPT-AFM framework.
In object tracking KLT method has been applied by Sundaram et.al\cite{Sun10} for multiple point tracking in a parallel environment. Chen et.al \cite{Che11} perform segmentation of video object and apply optical flow method to track each segments. Cui .et .al \cite{cui13}proposed a probabilistic fusion approach between low and high dimensional tracking approaches. Schwarz et.al\cite{Schw12} use optical flow in subsequent intensity image frames to get the motion information about the moving object body and apply graph based representation to track the entire object. Tsutsui \cite{tsu01} proposed an optical flow based object tracking method where multiple camera has been used instead of single camera. If the object is occluded by some other object different camera images helps the method to estimate the position of the object.In 2012 Liu et.al\cite{liu12} propose a fusion approach for object tracking of low and high dimensional tracking sampling information. In 2006, Brox et.al show a way of tracking 3D pose by optical method utilizing contour and flow based constraints in \cite{bro06}. 

 Aslani et.al \cite{Asl13} used optical flow together with some image processing method to estimate the position of the object in consecutive frames, and using that positional pixel values they track the whole object. Kale et.al\cite{Kal15} use optical flow to compute motion vector which provides an estimation of object position in consecutive frames. Though optical flow method has been applied extensively in object detection and tracking still there is no method which can extract perfect flow of data. Thus, the use of optical flow in object tracking is still a widely open problem\cite{Hus17}. In \cite{Wu13} Wu.et.al provide object tracking benchmark. In \cite{Wu13} Wu.et.al provide tracking results of some of the top performing object tracking algorithms: Visual Tracking via Adaptive Structural Local Sparse Appearance Model(ASLA) \cite{Jia12} Jia et.al use sparse representation to find possible match with target template with minimum reconstruction error, Beyond semi-supervised tracking: Tracking should be as simple as detection, but not simpler than recognition(BSBT) \cite{Sta09} Stalder et.al use multiple supervised and semi supervised classifier to perform the task of detection, recognition and tracking. Color-based probabilistic tracking[CPF] \cite{Per02}  Perez' et.al use Monte Carlo tracking method with particle filter. Exploiting the circulant structure of tracking-by-detection with kernels[CSK] \cite{Hen12} Henriques et.al use theory of circulant matrices with Fast Fourier Transformation to detect and track the moving object. Real-time compressive tracking[CT] \cite{Zha12} zhang et.al create an appearance model based on feature extracted from multi-scale image space and compute a sparse measurement matrix. Later using that sparse matrix they compress foreground and background targets, and perform tracking by using naive-Bayes classifier.
 
 In \cite{Mou17} Moudgil et.al provide a benchmark dataset for long duration video sequence which they name as 'Track Long and Prosper(TLP)'. This dataset is important because most tracking algorithms work well in short sequences but drastically failed on long challenging video sequence. This dataset contains 50 long time running video nearly 400 minutes. This \cite{Mou17} includes some of the recent tracking algorithms: Learning multi-domain convolutional neural networks for visual tracking \cite{Nam16} Nam et.al use convolutional Neural Network which is composed of different domain specific layers  which are trained to capture different parts of moving object to track. Fully-convolutional siamese networks for object tracking \cite{Ber16}  Bertinetto et.al create a fully-convolutional Siamese Network which is trained with ILSVRC15 dataset for video tracking. Crest: Convolutional residual learning for visual tracking \cite{Son17}  Song et.al reformulate Discriminative correlation filters as a one-layer convolutional neural network and apply residual learning to take appearance changes into consideration. Action-decision networks for visual tracking with deep reinforcement learning \cite{Yun17} Yun et.al propose a tracking algorithm which sequentially pursue actions learned by deep reinforcement learning. MEEM: robust tracking via multiple experts using entropy minimization \cite{Zha14} Zhang et.al propose multi-expert restoration method for problem of drifting of model in on line tracking by creating an expert ensemble where best expert is selected based on minimum entropy criteria to correct undesirable model updates.
 
 Bio-inspired based methods are effective tools for object tracking and are given extensive attention in past  few decades \cite{Sev12} Among other Bio-inspired methods like Genetic Algorithm (GA),Ant Colony optimization(ACO), Particle Swarm Optimization (PSO) emerges real fast because of its efficient, robust and quick convergence. Some of the earlier works successfully considered tracking problems using PSO. Particle Swarm Optimization is applied by Zheng et.al \cite{Zhe07}, \cite{Yuh08} on high dimensional feature space for searching optimal matching in Haar-Like features detected by a pre-defined classifier set. Xiaoqin et.al \cite{Xia08} calculate temporal continuity between two frames and use that information for swarm particle to fly and track that information. Vijay et.al \cite{Joh10} construct Human Body Model as a collection of truncated cones and numbering those cones and PSO cost function checks how well a pose matches with data taken from multiple cameras.  Multiple people tracking is considered by Chen et.al \cite{Chen11} where target object is modeled by feature vector and then PSO particles search the search space for optimal matching. Fakheredine shows \cite{Fak12} the use of multiple swarms for multiple parts of object during object tracking. Those swarms share information with each other to make tracking of object as a whole. Multiple object tracking is also considered by Chen-Chien et.al \cite{Hsu12} using PSO, construct a feature model using gray-level histogram and apply PSO particles to track the difference between gray level histogram information  of consecutive frames in a video sequence. Bogdan \cite{Bog13} represents an approach where object is represented by image template and a covariance matrix is formed on that. Using similarity measure PSO tracks the difference between the movement of object and target template \cite{Kim12}. We Compare our proposed approach with some other PSO based tracking algorithm: Multiple object tracking using particle swarm optimization \cite{Hsu12}  Hsu et.al first create a grey-level histogram feature model and then distribute PSO particles where target object used as fitness function. Real-Time Multiview Human Body Tracking using GPU-Accelerated PSO  \cite{Bog14} Boguslaw et. al show that movement is tracked by a 3D human model in the pose described by each particle and then rasterizing it in each particle’s 2D plane. Hierarchical Annealed Particle Swarm Optimization for Articulated Object Tracking \cite{Xua13} Xuan et.al show articulate object tracking by decomposing the search space into subspaces and then using particle swarms to optimize over these subspaces hierarchically.  Monocular Video Human Motion Tracking based on Hybrid PSO \cite{Ben14}  Ben shows tracking human pose in monocular video human motion by using hybrid PSO method. Object tracking using Particle Swarm Optimization and Earth mover's distance \cite{Xia17} Xia et.al use Particle Swarm Optimization (PSO) as the object localization method based on the Bayesian tracking framework.
 
 Though several approaches for object tracking are existing for past several decades but none of them consider dual tracking approach which is the major novelty of the present work.The proposed dual tracking approach for object tracking is very much robust for short video sequence under static background and variable background as well as long challenging video sequence under static background and variable background. Though an exhaustive survey on existing object tracking algorithms and a through review on deep learning approaches to object tracking are not within the scope of the present work; but in this particular context we like to make a critical appreciation on deep neural network(DNN) learning for object tracking. Through several experimental studies we reveal that deep learning approaches(based on Microsoft's ResNet, Google's Inception and Oxford's VGGNet etc.) for object tracking are good for known classes of object under fixed environments; but under variable background(in a time critical situation) with several uncertainties like unknown object,variation of illumination of light,noise, occlusion, sudden start and stop of moving object, shading etc. we observe that the deep learning method for object tracking don't produce satisfactory results and in many cases deep neural network(DNN) requires further training under the new environment and unknown objects \cite{Sim15}. In \cite{Sim15} Simpson states that "It is an embarrassing fact that while deep neural networks(DNN) are frequently compared to the brain, and even their performance found to be similar in specific static tasks, there remains a critical difference; DNN do not exhibit the fluid and dynamic learning of the brain but are static once trained. For example, to add a new class of data to a trained DNN it is necessary to add the respective new training data to the preexisting training data and re-train (probably from scratch) to account for the new class. By contrast, learning is essentially additive in the brain – if we want to learn a new thing, we do". Based on such observation \cite{Sim15} and a critical appraisal \cite{Mar18} and also based on our recent experimental studies on VGGNet we categories DNN approach as a representation of crystallized intelligence \cite{Hor68}, \cite{Cat63} of the network under learned or accumulated knowledge and has low capability of handling unknown environment specially under unknown objects. We suggest that for new environment with unknown objects DNN should have an added feature of fluid intelligence with some working memory which can handle novel or abstract problem solving environment \cite{Sch12}, \cite{Bla06}, \cite{Hor68}, \cite{Gra03}, \cite{Cat87}, \cite{Cat63}, \cite{Ash06}, \cite{Ack05}, \cite{Ack00}. However all such challenging issues and several other proposals \cite{Wol18}, \cite{Wol04}, \cite{Wol13}, \cite{Wol14}, \cite{Wolf14}, \cite{Wol16}, \cite{Wolf16}, \cite{Wol17}, \cite{Wolff16} should be throughly reevaluated before we come to any conclusion. Such issues should be separately considered elsewhere as an independent work.
 
 In this paper we essentially try to extract the merit of fusion between KLT tracker and PSO based tracker. Such fusion is considered to supplement each other in an intelligent fashion so that dual trackers become very simple, very robust and cost effective under variable background and static background and very much capable of handling the challenges of object tracking  which cannot be tackled by  DNN based tracking algorithms as stated above. The proposed dual tracking algorithm can successfully tracks object for short video sequence as well as long challenging video sequence. In case of unknown object , dual tracking approach simply needs to recalculate the dominant points on the contour of the unknown target object(objects) and no need to spend huge time to learn/train the unknown environment with unknown object form the beginning of tracking of the target object as we have seen in case of DNN based tracking algorithms. The KLT trackers based on optical flow concepts tracks the dominant points of the target objects from the first frame to last frame of the video sequence. Tracking of dominants points by KLT tracker is supplemented by swarm intelligence based PSO tracker from frame 2 to last frame. Swarm Intelligence based PSO (Particle Swarm Optimization) tracker basically tracks the boundary information of the target object from frame 2 to last frame. The flexibility of our approach is that it can be successfully applicable to variable background as well as static background. The basic tracking sequences of the propose dual tracking approach is as follows;
 
 In the first frame of the video sequence we obtain the dominant points of the target object and start tracking it by KLT tracker till the last frame. In frame 2 of the same video sequence the boundary of the target object is polygonally approximated for the first time. An environment of multiswarms is generated and an annular ring(strip) of swarms is formed within which the approximated ploygon is embedded.At frame -3 onwards the shape of the annular ring (strip) of the multiswarms changes simply because the shape of the dynamically generated polygon which, at each frame, continuously captures the boundary information of the target object, changes due to the movement of the target object which is a non-rigid body in general. The newly generated polygon embedded over the newly generated annular ring(strip) of the multiswarms is tracked from frame -3 by PSO tracker along with KLT tracker. The above process of dual tracking continues till the last frame with dynamic change of shape of the approximated polygon and the change of shape of the annular ring(strip) at each frame of the video sequence.
 
 In course of tracking if there is any loss of dominant points due to some sort of unpredictable disturbances then the tracking procedures by KLT and PSO are disturbed. In that case instead of recomputation of dominant points, we reinitialize the missing dominant points by some heuristic approach which essentially exploits the intelligence level of swarms. Similar to the reinitialization of the missing dominant points sometimes positions of the particles of the individual swarm may need to be reinitialized due to its distraction from the individual swarm by some process of disturbances. When all the swarms around the boundary of the target object reach the optimal solution a bounding box is generated around the target object based on particles final positions. This entire method of tracking uses only one feature which is basically dominant points on the contour of the target object and other information of the target object boundary is captured by the annular ring(strip) of multiswarms within which polygonally approximated target object is embedded. Note that the notion of using dominant points on the contour of a target object as good features for object tracking is basically derived from the concept of interest points as proposed by Shi et.al \cite{Shi94} and Tomasi et.al \cite{Tom91}.  In our approach, instead of searching for interest points of an object for tracking we directly compute dominant points on the contour of the target object to be tracked and thereby reduces the search complexity of KLT algorithm for object tracking. In sec 2.3.3 we experimentally demonstrate that the set of dominant points on the contour of the target object is basically a subset of interest points.Further note that the use of dominant points as good features for object tracking is an important and unique concept which is not used by classical KLT algorithm for object tracking. The robustness of the proposed dual tracking algorithm, under several existing challenges of object tracking as stated earlier, is verified  and established through several experimental studies on benchmark datasets \cite{Wu13}, \cite{Mou17}, \cite{Gei12}. Another specialty of the proposed dual tracking algorithm is its robustness under short video sequence as well as long challenging video sequence where most of the existing classical approaches fail \cite{Mou17}. Note that in the proposed dual tracking algorithm the KLT tracker for tracking the dominant points of the target object is continuously supplemented by PSO tracker from frame-2 to last frame. And due to embedding of the target object approximated by polygon in the annular ring(strip) of multiswarms,the target object is tightly captured throughout the tracking sequence by  the multiswarms environment and there is no loss of meaningful information about the target object during tracking. Hence the proposed dual tracking algorithm is inherently robust.
 
 There are several striking features of the proposed dual tracking algorithm.The overall performance of the proposed dual tracking algorithm, with respect to several benchmark datasets, are very much competitive and in most of the cases superior than the others.
 
 The paper is organized as follows: Section 2 discusses the basic concepts and tools and techniques required for dual tracking algorithm. Section 3 essentially deals with salient features of the proposed dual tracking algorithm.Section 4 pictorially describes the proposed dual tracking algorithm. Section 5 provides the pseducode and complexity analysis of the proposed dual tracking algorithm.Section 6 provides detail experimental studies on 3 benchmark datasets and also provides some analysis and performance measure of the proposed dual tracking algorithm. Section 7 provides Conclusion and future work.

\section{Dual approach for object tracking}
\label{sec:2}
\subsection{Basic concepts}
\label{sec:21}
In this paper we propose a dual approach for object tracking based on optical flow and swarm Intelligence. The optical flow based tracker i.e. KLT, tracks the dominant points of the target object from frame 1 to last frame; whereas swarm Intelligence based PSO (Particle Swarm Optimization) tracker simultaneously tracks the boundary information of the target object from frame 2 to last frame. This dual function of tracking makes the trackers very much robust with respect to the above stated problems.

In our approach, in the first frame of the video sequence we calculate the dominant points of the target object and start tracking it till the last frame. From frame 2 of the same video sequence the boundary information of the target object is captured by a dynamically generated polygon of the target object. The polygonal approximation of the target object at each frame is achieved by joining two consecutive dominant points on the target object by a straight line segment. In frame -2 of the same video sequence a group of particles is distributed randomly over the image search space. This particles form swarm over each line segment of the dynamically generated polygon of the target object. Formation of swarm on each line segment is based on the smallest distance of each particle from the individual line segment.

Thus, a multiswarms environment is formed and an annular ring(strip) of swarms is generated over which the dynamically generated polygon of the target object is embedded. If the target object is a closed digital curve then the annular ring of swarms is formed as shown in fig-(1); otherwise a strip of swarms is formed as shown in fig-(2).

\begin{figure}[h]
	\centering
	\includegraphics[width=0.5\textwidth]{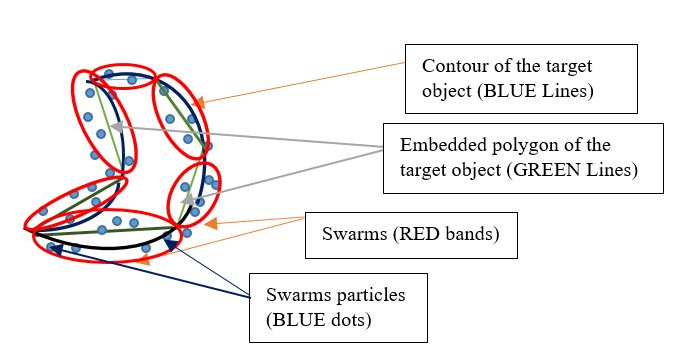}
	\caption{Annular ring of multiswarms.}
	\label{fig:1}
\end{figure}

\begin{figure}[h]
	\centering
	\includegraphics[width=0.5\textwidth]{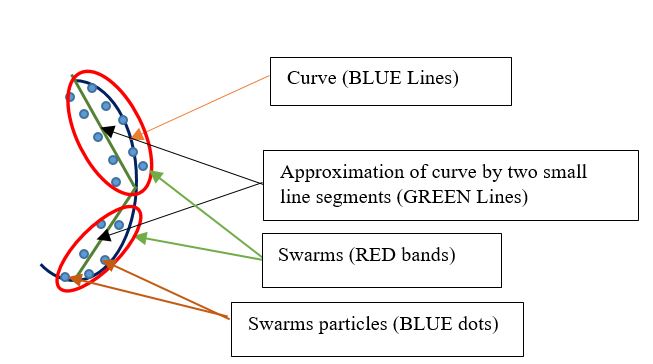}
	\caption{Strip of multiswarms.}
	\label{fig:2}
\end{figure}

The vertices's (dominant points) of the polygon are tracked by KLT tracker and the boundary information of the target object, which is approximated by dynamically generated polygon and which is embedded over the annular ring(strip) formed by multiswarms, is tracked by the pso tracker from frame -2 to last frame. At frame -3  the shape of the annular ring(strip) of the multiswarms changes simply because the shape of the dynamically generated polygon, which continuously captures the boundary information of the target object, changes due to the movement of the target object which is a non-rigid body in general. During the said process of shape change of the annular ring(strip) of multiswarms, the individual swarm of each small line segment further rearranges the position of the particles of each swarm to converge on the individual line segment of the newly generated polygon. During the said process of convergence, until all the particles of individual swarm over individual line segment successfully converges over all the line segments of the newly generated polygon, they (particles) update their velocity and position based on previous local best and global best position. Thus local best and global best positions are further updated. Again the newly generated polygon embedded over the newly generated annular ring(strip) of the multiswarms is tracked from frame -3 by PSO tracker along with KLT tracker. The above process of dual tracking continues till the last frame with dynamic change of shape of the polygon and the change of shape of the annular ring(strip) at each frame of the video sequence. Thus the dual tracking approach for object tracking tracks the dominant points on the contour of the target object and simultaneously tracks the tightly captured and embedded approximated polygon of the target object. The basic purpose of this dual tracking approach is that during tracking the multiswarms environment within which the approximated polygon is embedded continuously supplement the KLT tracker for tracking the dominant points  from frame-2 to last frame. As the polygonally approximated target object is embedded and tightly captured within the frame of multiswarms ring(strips) so under any kind of environmental disturbances as stated earlier the tracking of the target object is not lost in the midway of any video sequence of tracking. Another specialty and uniqueness of this dual tracking approach is that it very successfully tracks the long challenging video sequences where many classical approaches for tracking drastically fails. This achievement of successful tracking of long challenging video sequence is mainly due to the fact that the approximated polygonal version of the target object is embedded and tightly captured in a multiswarms environment. And there is a very little possibility that the target object is lost during tracking  in a long challenging video sequence.

In course of tracking if there is any loss of dominant point (points) due to some environmental disturbances then the tracking procedures by KLT and PSO are disturbed. In that case instead of recomputation of dominant point (points), we reinitialize the missing dominant point (points) by some heuristic approach which essentially exploits the intelligence level of swarms. Similar to the reinitialization of the missing dominant point (points), particles of the individual swarm over individual line segment may require reinitialization during convergence process, which starts from frame -3 till the end of the last frame. If it is detected, during the said convergence process, a particular particle(particles) of an individual swarm over individual line segment diverges ( instead of converges) from its global best position ( even after several iteration of convergence) then the position(positions) of that particular particle(particles) is (are) reinitialized to a position(positions) for convergence over the line segment of the corresponding swarm from where the particle(particles) is (are) displaced to an undesirable position. After successful convergence of all particles over individual swarm of each line segment of the polygon a bounding box around the target object is formed based on a new concept of PSO-based bounding box generation algorithm.Note that, as stated earlier,at frame-2 a group of particles are randomly distributed over the image search space. These particles essentially take part in formation of swarms on individual line segments of the dynamically generated polygon of the target object. The population of particles is not fixed. It depends upon the need of the problem and basically a heuristic parameter in nature \cite{Roh11}, \cite{Xue15}. If the population of the particles at frame-2 is very large the computational complexity of the entire algorithm may increase. Keeping this in mind we have to select the population of particles at frame 2.The flexibility of our approach for dual tracking is that it can be successfully applicable to variable background as well as static background. The tools and techniques used for implementing the basic concepts of dual tracking are discussed in the following --

\subsection{Dominant Point Detection}
\label{sec:22}
For the detection of the dominant point on the contour of the target object we use the methods \cite{Ray92}, \cite{Ray13} and \cite{Wu003} and \cite{Wu13}. We first perform contour tracking of the target object to find the Chain Code based on Freeman's Chain Code \cite{Fre61}. \textit{Freeman Chain code} gives us list of pixels around object body. Among those pixels we eliminate linear points(pixels), as those points(pixels) do not provide us any significant curvature information. For elimination of linear points(pixels) we consider the following rule --

\begin{equation}
\textrm{if}\ C_{i-1} = C_{i}\textrm{ then point}\ P_{i} \textrm{ is a linear point,}
\end{equation}
where $C_{i-1}$  is the previous chain code value and $C_i$ is the current one, on the point $P_i$.

After excluding those linear points(pixels) rest of the points(pixels) are called breakpoints, which are candidates for dominant points. We have to consider the region of support of only for those breakpoints. We calculate the length of support of each breakpoint. Rather considering all breakpoints at once we collect them as a group of 10 for variable background and group of 5 for static background. The number of breakpoints in a group is decided based on which background we perform the tracking. Normally on variable background object shape changes fast. Hence we need the curvature of the object body smaller so that large number of breakpoints are close to each other. That’s why we chose large number of breakpoints, compare to static background where the object is more stable and we can use much longer curvature. Therefore less number of breakpoints suffices for dominant point calculation.

For each group of breakpoints we calculate k-Cosine values for each of them and apply the following rule -
Let us start with k =1 to form a group. Increase the value of k by 1 until we reach all breakpoints on that group.

\begin{equation}
\begin{split}
k_{i} = k \textrm{ if }cos_{ik} = max\{cos_{ij}|j = K_{min}...K_{max}\}, \\
\textrm{ for } j = 1,2,3...n.
\end{split}
\end{equation}

We chose dominant point as those points which are \textit{\textbf{max k$-$Cosine}} values, i.e.\\
\begin{equation}
D_i = max (cos(P_i)).
\end{equation}

\paragraph{Dominant Point Calculation : }
Thus the entire procedure for calculating dominant points can be summarized as follows;
\begin{itemize}
	\item Use Freeman Chain Code for performing contour tracking.Get those pixels and store them in a file.
	\item Form the stored pixels eliminate linear points using equation (1). Save them in a file and call breakpoints.
	\item Perform K –Cosine for each of the breakpoints using Eq-(2).
	\item Select those points as dominant points which has max k-cosine values and collect a set of Dominant points as per Eq-(3).
\end{itemize}

\subsection{Tracking of Dominant point(points) by KLT}
\label{sec:23}
\subsubsection{Feature selection}
\label{sec:231}
Before any tracking of moving object the most fundamental step is the selection of “trackable" features. For the present problem we consider sparse optical flow method based on KLT algorithm. First we have to determine the parameters to find out good features. According to Tomasi and Kanade\cite{Tom91} \textit{'a single pixel cannot be tracked until it has s a very distinctive brightness with respect to all of its neighbors'}. Hence, they prefer a “Window” of pixels which should contain sufficient texture. By “texture” we mean a group of neighboring pixels (window of pixels) which shows significant variation or changes of intensity or brightness between consecutive frames. Areas with a varying texture pattern are mostly unique in an image, while uniform or linear intensity areas are often common and not unique. Based on these guideline we proceed as follows;

\subsubsection{Selecting Dominant point(points) as good feature}
\label{sec:232}
The main reason for choosing dominant point as a trackable feature is that by definition \cite{Ray13} dominant point itself holds maximum curvature information on the contour of a target object. So quite obviously a window centered at dominant point should always give us enough texture for tracking from one frame to another. The area of such window can vary, depending on the number of features. This dominant point act as “interest point” which captures maximal local intensity information.
Every basic KLT algorithm starts with finding corners or interest points satisfying the equation.
\cite{Shi94} -

\begin{equation}
Min(\lambda_1,\lambda_2) = \lambda
\end{equation}
where, $\lambda_1,\lambda_2$ are two eigenvalues and $\lambda$ is a predefined threshold. Rather applying a separate algorithm for finding good interest points which satisfy the above equation-(4), we consider dominant point as our interest point. Let us define the image gradient as follows --
\begin{equation}
%\[
G=\left [
\begin{tabular}{c}
$G_x$\\
$G_y$
\end{tabular}
\right ].
%\]
\end{equation}
We, consider the product of gradient and its transpose as follows --
\begin{equation}
%\[
GG^T=\left [
\begin{tabular}{p{3em} c c}
$G_x^2$ & $G_x$$G_y$\\
$G_x$$G_y$ &$G_y^2$
\end{tabular}
\right ].
%\]
\end{equation}
If we integrate the matrix defined above over the area W(selected window),we get.
\begin{equation}
Z=\iint_W \left [
\begin{tabular}{p{3em} c c}
$G_x^2$ & $G_x$$G_y$\\
$G_x$$G_y$ &$G_y^2$
\end{tabular}
\right ] W dx.
\end{equation}

Z is a 2x2 matrix containing texture information along X and Y axis. Analyzing the eigenvalues of the matrix Z we get the W, which is window of pixels that are trackable.The equation for Z forms an intricate part of the Kanade-Lucas-Tomasi tracking algorithm. It is necessary to establish a minimum threshold for the value of the eigenvalues. If the two eigen values of Z are $\lambda_1$ and $\lambda_2$, we accept a window which satisfies equation–(4).

\begin{figure}[h]
	\centering
	\includegraphics[width=0.5\textwidth]{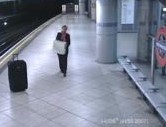}
	\caption{original image; no dominant point is selected so far.}
	\label{fig:3}
\end{figure}

\begin{figure}[h]
	\centering
	\subfloat[image gradient according X- axis.]{
		\resizebox*{7cm}{!}{\includegraphics{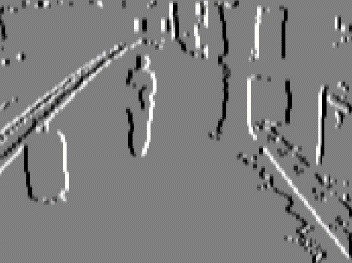}}}\hspace{4pt}
	\subfloat[image gradient according Y- axis.]{
		\resizebox*{7cm}{!}{\includegraphics{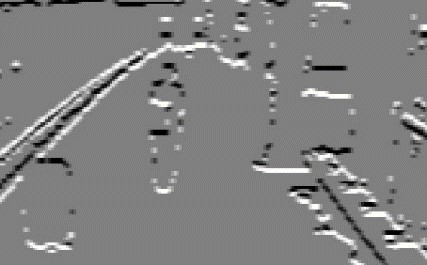}}}\hspace{4pt}
	\caption{image gradient according X and Y- axis from left to right.}
	\label{fig:4}
\end{figure}

\begin{figure}[h]
	\centering
	\includegraphics[width=0.5\textwidth]{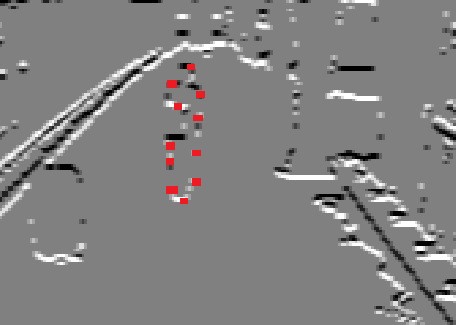}
	\caption{RED dots indicate the pixels which qualify the equation-(4).}
	\label{fig:5}
\end{figure}

\subsubsection{Dominant points as subset of interest points}
\label{sec:233}
In section 2.2 we state that dominant point holds maximum curvature information on the contour of a target object and provides enough texture for tracking. In this subsection we further clarify this concept through a simple experiment, as example, that dominant points are the subset of interest points which are the key elements of KLT tracking algorithm. In Figure-(5) the set of RED dots are the interest points as per equation -(4) and our chosen dominant points for target object are taken from this set of RED dots as a subset.

In figure-(3) we show original image and in figure-(4) we show image gradient in x axis and y axis. Figure-(5) is the results of the feature points which satisfy the above equation-(4).Experimentally we obtained that the calculated dominant points using equation -(3) \cite{Ray92} is a subset of the interest points of a selected window of feature points as stated above. So we can move to the next step of the KLT tracking algorithm by considering dominant points as our interest points which we do not have to search for \cite{Shi94}

\subsubsection{Concepts of tracking dominant points by KLT}
\label{sec:234}
The basic notion of tracking by KLT can be explained by looking at two images in an image sequence. Let us assume that the first image is captured at time t and the second image is captured at time t + $\uptau$ . It  is important to keep in mind that the incremental time $\uptau$ depends on the frame rate of the video camera and should be as small as possible. An image can be represented as function of variables x and y. If we define a window in an image taken at time t+$\uptau$ as  I(x,y,t+$\uptau$). The basic assumption of the KLT tracking algorithm is;
\begin{equation}
I(x,y,t+\uptau) = I(x-\Delta x,y-\Delta y,t)
\end{equation}

From equation -(8) it is clear that every point in the second window can by obtained by shifting every point in the first window by an amount ($\Delta$x, $\Delta$y). This amount can be defined as the displacement d = ($\Delta$x, $\Delta$y) and the main goal of tracking is to calculate d.

\subsubsection{Calculation Feature displacement}
\label{sec:235}
Now we have basic information to solve the displacement d mentioned above. The solution is explained in \cite{Bir97}. According to \cite{Bir97} --
we can calculate displacement d from from image frame I to image frame J.Thus we obtain-
\begin{equation}
\epsilon = \iint_W\Big[
J(x + \frac{d}{2}) - I(x - \frac{d}{2})
\Big]^2 W(x) dx
\end{equation}
where $x =[x \quad y]^T$ , the displacement $d =[dx \quad dy]^T$ , and the weighting function w(x) is usually set to constant 1. Now according to Taylors series expansion of J about a point $p(x,y)T$, truncated to the linear term is $-$
\begin{equation}
J(\epsilon)\approx J(p) + (\epsilon_x - a_x)\frac{\delta j}{\delta x}(a) + (\epsilon_y - a_y)\frac{\delta j}{\delta y}(a)
\end{equation}
where, $\epsilon$ = $[\delta x \quad \delta y]^T$.
Following the derivation, we let (x + $\frac{d}{2}$) = $\epsilon$ .To get the final derivation,
\begin{equation}
\begin{split}
\frac{\delta\epsilon}{\delta d} = 2 \iint \Big[
J(x + \frac{d}{2}) - I(x - \frac{d}{2})\Big]
\Big[
\frac{\delta J(x + \frac{d}{2})}{\delta d} - \frac{\delta I(x - \frac{d}{2})}{\delta d}
\Big] W(x) dx
\end{split}
\end{equation} 
In continuation of equation-(11) we calculate,

\begin{equation}
\frac{\delta\epsilon}{\delta d} \approx \iint_W \Big[
J(x) - I(x) + g^Td
\Big]g(x)W(x)dx
\end{equation}
where, 
$ g = \Big[
\frac{\delta}{\delta x}(\frac{I+J}{2}) \quad \frac{\delta}{\delta y}(\frac{I+J}{2})
\Big]^T.$
To calculate the displacement d , we need to set the derivative 0.
\begin{equation}
\frac{\delta\epsilon}{\delta d} = 0.
\end{equation}
Solving further, we get a simplified equation -
\begin{equation}
Zd = e.
\end{equation}
where, Z is the 2x2 matrix :
$ Z =\iint_W g(x)g^T(x)W(x)dx$ and e is the 2x1 vector:
$e =\iint_W\big[ I(x) - J(x)\big]g(x)w(x)dx$. So the displacement d is the solution of equation-(14).

\subsubsection{KLT algorithm} 
\label{sec:236}
We summaries the KLT algorithm as follows -\\
Step 1: Find the dominant points which satisfy min($\lambda_1,\lambda_2) > \lambda$(see equation -(4).\\
Step 2: For each dominant point compute displacement to next frame using the Lucas-Kanade method (see equation -(14)).\\
Step 3: Store displacement of each dominant point, update the position of the dominant point.\\
Step 4: Go to step 2 until all dominant points are exhausted.

\subsection{Particles Swarm Optimization (PSO) method for tracking.}
\label{sec:24}
In 1995 James Kennedy and Russell Eberhart proposed an evolutionary algorithm that creates a ripple among Bio-inspired algorithms. This particular algorithm is called Particle Swarm Optimization (PSO)\cite{Ebe95}. In a simple term it is a method of optimization for continuous non-linear function. This method is influenced by swarming theory form biological world like fish schooling, bird swarming etc\cite{Ahm12}.

PSO is effectively applied to the problems in which each solution of that problem can be considered as \textit{a point} in a solution space. Each point in the solution space is called one \textit{particle}. Analogically suppose there is a food source and a swarm of birds tries to reach that food source. Every bird, tries by its own choice to reach there. Whoever is reached. or nearly reached to that food source share that information with other birds who are close neighbor.As a ripple in water information flows among entire swarm of birds and every bird synchronously update their velocity and position. If it gets better position in terms of nearest position to the food source. As a result after certain period of time entire swarm eventually gathers to the food source. Every solution considered as particle computes its value based on some cost function, until it satisfies certain criterion known as stopping condition. It keeps updating its velocity and position, provided its neighbor has better solution.

Position and Velocity are two associated terms in Particle Swarm Optimization. Position of every particle is calculated by particle’s own velocity. Let $X_{i}$(t) denote position of particle i in the search space at time t. Position updation formula is as follows --

\begin{equation}
X_{i}(t+1) = X_{i}(t) + V_{i}(t+1)
\end{equation}
where,

$V_{i}(t+1)$ is the velocity of particle i at time (t+1), which is computed based on this following formula-
\begin{equation}
\begin{split}
V_{i}(t) = V_{i}(t-1)+C_{1}.R_{1}(P_{LB}(t) - X_{i}(t-1)) + C_{2}. R_{2}\\(P_{GB}(t) - X_{i}(t-1))
\end{split}
\end{equation} 
where,
$C_{1}$,$C_{2}$ represent the relative influence on social and cognitive components respectively. They are also known as \textit{learning rates} and are often set to same constants value, to give each component equal weight.

$R_{1}$,$R_{2}$ = random values associated with learning rate components to give more robustness.

$P_{LB}$ = Particle Local Best position $-$ it is the historically best position of the $i^{th}$ particle achieved so far.

$P_{GB}$ = Particle Global Best position $-$ it is the historically best position of the entire swarm. Which is basically the position of a particle which achieves closest solution.
Equation (16) is Kennedy and Eberhart’s original idea. After that lot of different researches have been going on.Based on those researches a remarkable idea comes up \cite{Shi98}. In \cite{Shi98} Shi and Eberhart add a a new factor called ``inertia weight'' or ``w''. After addition of inertia weight the Eq (14) becomes as follows --
\begin{equation}
\begin{split}
V_{i}(t) = w * V_{i}(t-1)+C_{1}.R_{1}(P_{LB}(t) - X_{i}(t-1)) + C_{2}. R_{2}\\(P_{GB}(t) - X_{i}(t-1))
\end{split}
\end{equation} 
This inertia weight helps to balance local and global search abilities. Small weight means local search and larger weight means global search \cite{Car01}. Pseudo code of the basic PSO algorithm  is given in appendix.

In this paper the PSO based tracker tracks the dynamically approximated polygon of the target object and continuously supplements the tracking of the dominant points of the target object by KLT.

\subsubsection{Setting PSO parameters and Initialization}
\label{sec:241}
Because of dynamic nature, setting PSO parameters to right value is a crucial task. Below we discuss some of the major parameters.

\begin{itemize}
	\item{\textbf{Multiswarms} } - In the proposed dual tracking algorithm one tracker is PSO based approach. In the basic concept of section 2.1 we have clearly explain a key feature of dual tracking algorithm is ring(strip) of multi swarms within which the approximated polygonal representation is embedded during tracking. Number of swarms are decided by number of dominant points of the target object. If we have D number of dominant points of a target object the number of line segments which polygonally approximate the said target object is equal to (D - 1). Thus there will be (D - 1) number of swarms. During tracking, due to several disturbances as stated earlier the dominant point(points) of the target object may be lost at the midway of tracking and thereby some of the particle of the swarm which is based on that dominant point will be also distracted. In such cases as mentioned in section 3.1 and section 3.2 the algorithm will automatically reinitialize the lost dominant points and the lost particle of the swarms. During experiment of tracking we have seen in worst case approximately 10\% percent of the particle including dominant points need to be reinitialized.  
	
\item{\textbf{Population of particles }} - We initialize the population of particles needed for construction of swarm around individual line segment of the approximated polygon of the target object. It is chosen  heuristically depending upon the need of the application. In case of object tracking under static background the particle population is 25 and object tracking under variable background the particle population  is 33. It is obvious that if we increase the population size the computational complexity of the PSO tracker will be increased. It is also obvious that more the length of the line segment more the particles will converge on that to form a swarm as per the PSO algorithm.

\item{\textbf{Position and Velocity initialization}} - According to PSO methodology we need to initialize the position and velocity of every particle of the swarm. This position of particle for each swarm will be inside the search space and randomly defined. In our case we first consider the range of the image space within which the target object are lying. If the image space is represented by [x,y] range for each frame then we first select some values for $V_x$ $V_y$ as velocity component in x-direction and y-direction to be \cite{Shi01};
		\begin{center}
		1 $\leq$ $V_x,V_y$ $\leq$3
	\end{center}
	where $V_x$ and $V_y$ denotes velocity towards X and Y direction.\\
	Velocity signifies how far a single particle will jump, as we are working on image pixels. It cannot be negative value or fractional value and also setting high value is not a practical approach as the particles work in close vicinity on the dominant points.
	
	\item \textbf{Local best value of Particle i ($P_{lbest_{i}}$)} -local best value of an individual particle in a swarm indicates its current best position it achieved to converge on the target line segment between two consecutive dominant points. We initialize each particle's Plbest value with its initial position inside the search space [x,y] which is randomly defined at the very beginning as stated above. Latter it will be modified according to the Plbest updating rule.
	
	\item \textbf{Global best value of Particle i ($P_{gbest_{i}}$)} - In a particular swarm, the particle which holds the best position such as close to the line segment between two consecutive dominant points is considered as global particle and its position is $P_{gbest_{i}}$. Each particle first compute the perpendicular distance from line segment connected by dominant points. The particle which hold minimum distance considered as $P_{gbest_{i}}$.
	
		\begin{equation}
	P_{gbest_{i}} = min
	\begin{cases}
	\sqrt{(X_{D1} - X_i)^2 + (Y_{D1} - Y_i)^2} \\
	\sqrt{(X_{D2} - X_i)^2 + (Y_{D2} - Y_i)^2}
	\end{cases}
	\end{equation}
	
	where ($X_{D1},Y_{D1}$) is the position of the 1st dominant point and ($X_{D2},Y_{D2}$) position of the 2nd dominant point and $X_{i},Y_{i}$ is the coordinate of the $i^{th}$ particle.
	
	\item \textbf{Initialization of w, $C_{1},C_{2},R_{1},R_{2}$ values.} - Earlier we define the meaning of these terms in equation -(16) and equation-(17). Initialization of these variables based entirely on application . In this paper after some experiment we choose w = 0.3 , $C_1$ = 0.1, $C_2$ = 0.1. $R_1$ and $R_2$ are set to some integer between 1 to 3. The convergence of the PSO algorithm is based on these parameters and we are basically guided by the information provided in \cite{Shi01}.
\end{itemize}

\subsubsection{Polygonal approximation of the target object}
\label{242}
For polygonal approximation of the target object we draw small line segments between two consecutive dominant points of the target object. Let us  consider two dominant points  $D_1$ and $D_2$  which are calculated using equation-(3). The path between this two points is a small line segments joining the said two points. There could be infinitely many curves( not straight line segments) that may pass through the said  two dominant points, but in this paper we consider Euclidean Distances between the two said points.\\ 
In Cartesian coordinate, $D_1 (X_1,Y_1)$ and $D_2(X_2,Y_2)$ are the two points in Euclidean space. The distance between this two points is calculated as follows -
\begin{equation}
\overline{D_1D_2} =\sqrt{(X_2 - X_1)^2 + (Y_2 - Y_1)^2}.
\end{equation}
In the following we illustrate this phenomenon using an arbitrary curve as shown in figure-(6):.
In the following figure-(6) , we have an arbitrary curve C as stated above, which contains dominant points like $D_1,D_2, D_3, D_4$. The line segments $L_1, L_2$  and $L_3$  passing through $D_1D_2$ , $D_2D_3$ and $D_3D_4$ respectively.

Though it is not exactly the curve connecting dominant points $D_1D_2$ , $D_2D_3$ or $D_3D_4$ but as shown in the figure it serves the purpose of  approximately representing the contour(boundary) of the curve as shown in fig-1. Thus we obtain polygonal approximation of the arbitrary chosen curve C as stated above. As we are not detecting or tracking the exact contour of the target object, we focus only on moving area of the target object approximated by polygon, so polygonal approximation of the target object does not produce any serious threat for tracking. It is always possible to construct the exact curvature between two consecutive dominant points of the target object. But such construction of the curvature is always time consuming and does not really improve the tracking result. This can be a scope for future work.

\begin{figure}[h]
	\centering
	\label{fig:6}
	\includegraphics[width=0.5\textwidth]{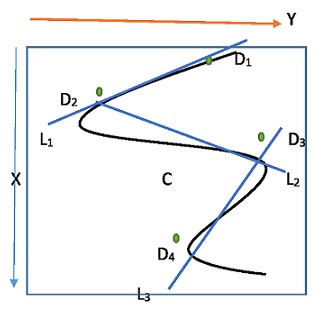}
	\caption{Diagram shows approximate Curvature calculation using Pythagorean formula.}
\end{figure}

\subsubsection{Fitness Function for PSO tracker}
\label{243}
Every PSO model is based on some cost function. Each particle of the swarm computes that fitness function in each iteration to confirm whether it converges to the final solution or not. In this paper, our cost function is the  distance of the particle i to the small line segment which is a part of the approximated polygon of the arbitrary curve C. 

\begin{figure}[h]
	\centering
	\label{fig:7}
	\includegraphics[width=0.5\textwidth]{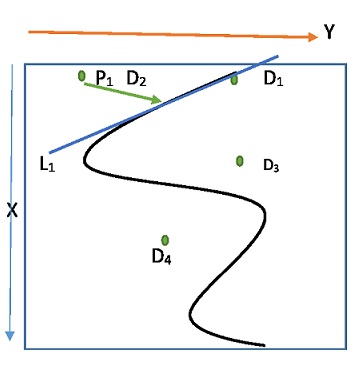}
	\caption{Generation of Fitness function.}
\end{figure}

Based on figure-(6), we draw figure-(7). Here we have a particle $P_1$ and the small line  is $L_1$. We compute the distance from the point $P_1$ to the line $L_1$.\\
As $L_1$ passes through two dominant points $D_1(X_1,Y_1)$ and $D_2(X_2,Y_2)$ then the distance from the point $P_1(X_0, Y_0)$ is --

\begin{equation}
Dist (D_1,D_2,P_1) = \nonumber
\end{equation}
\begin{align}
\frac{|(Y_2 - Y_1) * X_0 - (X_2 - X_1)*Y_0 + X_2 * Y_1 - Y_2 * X_1|} {\sqrt{(Y_2 - Y_1)^2 + (X_2 - X_1)^2}}.
\end{align}

The denominator is the length between  $D_1$  and $D_2$. Numerator is the twice the area of triangle with its vertices's at 3 points $D_1, D_2, P_1$.
For Every particle we compute the distance from the particle to the small line joining two dominant points as stated above and if the distance is in the acceptable range iteration stops else this procedure continues.

In our code , in each iteration we keep on running unless all the particles achieve desired threshold values, that means all the particles are as much as close to the line segment joining two dominant points.
So it is not just a single run of PSO algorithm rather an continuous loop until all the particles achieve threshold value.

One obvious question is why we choose line curve as a fitness function? We use a fitness function that is most suitable for tracking a curve. Though, we can use other fitness function like Beizer Curve or Hermite Curve but we think that when we are tracking a line segment joining two dominant points, using a line function as fitness function is most appropriate and easy to implement. We need approximation of the curvature not exact contour, so we stay away from complex fitness function and chose line function.
\subsubsection{Formation of Multiswarms}
\label{244}
Once the task of constructing the polygon of the target object is completed, for the first time,at frame 2 of the video sequence, we distribute particles over the entire image space. Note that the the population of the particles is a heuristic parameter which depends on the need of the problem and which has several options as stated in \cite{Dem12}, \cite{Mel09}, \cite{Jia07}. These said particles form swarm over each small line segment of the polygon according to the smallest value of the fitness function.

For the first time each particle of frame 2 measures its perpendicular distance from each small line segment and chooses the particular line segment as the line over which it will lie to form a swarm. Thus all particles of the image space of frame 2 are distributed over the small line segments of the approximated polygon of the target object and form a multiswarms scenario at frame-2 of the video sequence. These multiswarms scenario is nothing but an annular ring(strip) of swarms within which the approximated polygon of the target object is embedded(see figure-1 and 2 of section 2.1).

Dual tracking of the target object starts from frame -2 . The vertices's of the polygon which are essentially the dominant points of the target object and which are computed at frame 1 of the video sequence are tracked at frame -2 where these vertices's are embedded in the annular ring(strip) of the multiswarms as dominant points of the approximated polygon of the target object. These vertices's (dominant points) are tracked by KLT. KLT tracks the dominant point of the target object from frame 1 to the last frame of the video sequence. Whereas the entire target object which is approximated by polygon and embedded in the multiswarms environment is tracked by the PSO tracker. PSO tracks the approximated polygon of the target object embedded in the multiswarms from frame 2 to last frame.

\subsubsection{$p_{lbest},p_{gbest}$ Updation and Reshaping of the annular ring of the multiswarms}
When the dual trackers arrive at frame-3 of the video sequence, the shape of the polygon is automatically changed due to the movement of the target object which is in general non rigid in nature. In case of rigid object the shape of the polygon of the target object remains same. Once the shape of the polygon changes at frame 3 of video sequence the particles inside a swarm are redistributed on the small line segments of the changed polygon as per the built in function $p_{lbest},p_{gbest}$ function. Until all the particles inside a swarm successfully converge on the small line segment of the changed polygon, they (particles) keep updating their velocity and position using formula-(16) and (15) respectively.

$P_{gbest}$ and $P_{lbest}$ are updated as --

\begin{equation}
New\_P_{gbest}(P_i) = {min (PerDist (D_1,D_2,P_i)) \forall i }
\end{equation}
$New\_P_{lbest}(P_i)$ = 
\begin{equation}
\begin{cases} 
PerDist(D_1,D_2,P_i),&\textit{PerDist($D_1,D_2,P_i$) $<$}\\
&\textit{Previous PerDist($D_1,D_2,P_i$)}.\\
Previous PerDist (D_1,D_2,P_i), & \textit{Otherwise}.
\end{cases}
\end{equation}

\subsubsection{Reinitialization of the particle of the individual swarm}
\label{246}
At the time of updating the position the particles of the individual swarms, instead of converging over the small line segments of the changed polygon, the particle(particles) may be distracted from the said line segments to a far away distance even after several iteration of updation. In that case we need to reinitialize the particle(particles). Reinitializing particles over entire image space certainly feasible but not a practical idea. For further illustration see section-3.

\subsection{Bounding Box formulation}
\label{25}
To identify tracked target object usually a rectangular bounding box is utilized. There are some pre-defined algorithms exist for this purpose, but here we design our own bounding box based on PSO particle position which will best suite our object tracking algorithm.

The main idea is whenever all particles in all swarms successfully converge for a particular image frame we find p number of particles which have smallest X – direction and smallest Y-Direction. These particle are close to (0,0) in our image space. These p values can be the first 10 particles with minimum values in X- direction and Y-directions. This choice of number of particles entirely depends on application. As per our experimental experience this number of particles should lie from 10 to 20 particles with smallest X,Y direction. We take an average of these p- points, which is the starting point for bounding box formation.
Let us consider a particle q, which is  calculated as follows --

\begin{equation}
q = \Bigg\lceil{{\frac{(p_1 + p_2 + ... + p_p)}{p}}}\Bigg\rceil
\end{equation}

\begin{figure}[h]
	\centering
	\label{fig:8}
	\includegraphics[width=0.5\textwidth]{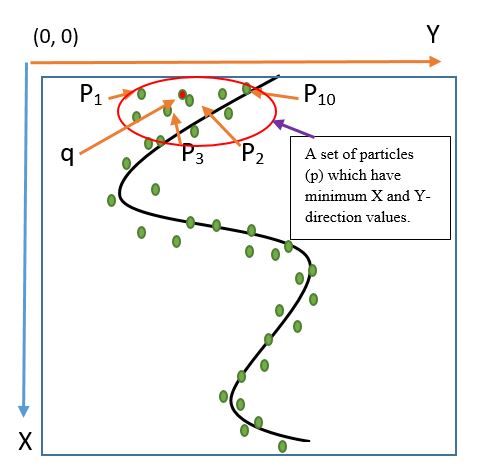}
	\caption{Convergence of particles towards object boundary for construction of bounding box.}
	\end{figure}

Figure-8 represent the point q.  Now we compute the Length and Breadth of the bounding box. Length is the vertical line and they are calculated as follows -

First select l number of particles which have maximum X-direction values and minimum  Y-direction values. These l values can be the first 10 particles which have values maximum  in X-direction and minimum in Y-direction. This number of points depends on designer choice and application. According to our experience it is effective if we take first 10 to 20 particles which are maximum in X  direction and minimum in Y direction. Thus we get as follows - 

\begin{equation}
len = \frac{\sum_{i=1}^{l} l_{i}}{l}  \forall i=1,2,3...l.
\end{equation}
Length is the euclidean distance from point q($q_{x},q_{y}$) to point $len(l_{x},l_{y}$) as shown below.

\begin{equation}
Length(L) = \sqrt{(q_x - l_x)^2 + (q_y - l_y)^2}.
\end{equation}
Similarly, for breadth calculation, first select b number of particle which have minimum X-direction values and maximum  Y-direction values.

\begin{equation}
bre = \frac{\sum_{i=1}^{b} b_{i}}{b}  \forall i=1,2,3...b.
\end{equation}

Breadth is the euclidean distance from point q($q_{x},q_{y}$) to point $bre(b_{x},b_{y}$) as follows.

\begin{equation}
Breadth(B) = \sqrt{(q_x - b_x)^2 + (q_y - b_y)^2}.
\end{equation}

Once we get the 3 parameters; length(L), breadth(B) and starting point q, using equation-(28) we construct the bounding box as shown in Fig-9.

\begin{figure}[h]
	\centering
	\label{fig:9}
	\includegraphics[width=0.5\textwidth]{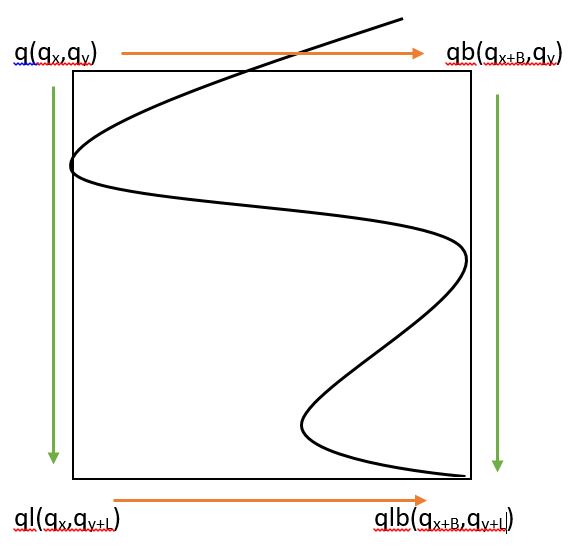}
	\caption{Representation of the bounding Box.}
\end{figure}

\begin{equation}
\begin{rcases}
&\text{$q=(q_x,q_y)$}\\
& \text{$ql=(q_x,q_{y+L})$} \\
& \text{$qb=(q_{x+B},q_y)$} \\
& \text{$qlb=(q_{x+B},q_{y+L})$}
\end{rcases}
\text{four boundary position formula}.
\end{equation}

\section{Salient features of the proposed algorithm.}
\label{sec:3}
\subsection{Re-initialization of missing dominant points}
\label{sec:31}
Due to background clutter, occlusion, illumination Variation, low resolution and scale variation of various video sequences, change of image background occurs frequently. So the optical flow method based Kanade--Lucas--Tomasi(KLT) tracker which is basically a point tracker is unable to track a single point throughout the video duration. Hence the proposed algorithm is developed based on the fusion between optical flow and swarm intelligence. After the first frame of tracking using KLT, the PSO provides a continuous support to capture the overall information including the dominant points of the object by automatic generation of polygon of the object to be tracked where the vertices's of the polygon is basically the dominant points of the object being tracked.This polygon is automatically updated with the moving object from frame to frame. With the movements of the object being tracked the shape of the object(usually non rigid) changes which is continuously updated by the newly generated polygon of the object at each frame.Thus a total information, in an approximated sense, is provided to the tracking algorithm by the dual function of optical flow and swarm intelligence.

We track dominant points using KLT. As the video sequence changes a lot there is a very high probability that KLT tracker may loose some of these dominant points in course of its tracking. During tracking, if the tracking of a dominant point is disturbed then the particle(particles) of the corresponding swarm is(are) also distracted and PSO tracker may failed to track. To avoid this situation we propose reinitialization of missing dominant points.

\subsubsection{Pictorial Illustration of Reinitialization of missing dominant points}
\label{Sec:311}
We explain reinitialization process of missing Dominant points using an example. Let’s consider a curvature C whose start (s) and end (e) points are dominant points. We track these dominant point using KLT method.In the following figure-(10) there are two dominant points (s and e) which are  marked as RED. These two points are tracked by KLT tracking algorithm and all yellow color points are swarm particles which spread over the line joining between two dominant points.
\begin{figure}[h]
	\centering
	\label{fig:10}
	\includegraphics[width=0.6\textwidth]{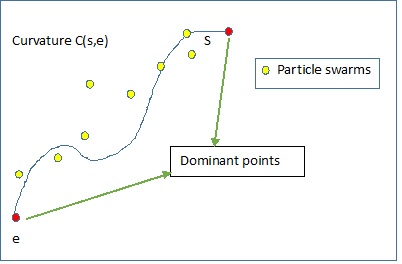}
	\caption{Initial curvature C(s,e) with two dominant points (RED dots) tracked by KLT and yellow dots represent the swarm particles.}
\end{figure}
If we consider that the curved  object which is being tracked is moving from left to right then due to various reasons stated above KLT may loose tracking of one of the dominant points as shown in the figure-(11).
\begin{figure}[h]
	\centering
	\label{fig:11}
	\includegraphics[width=0.6\textwidth]{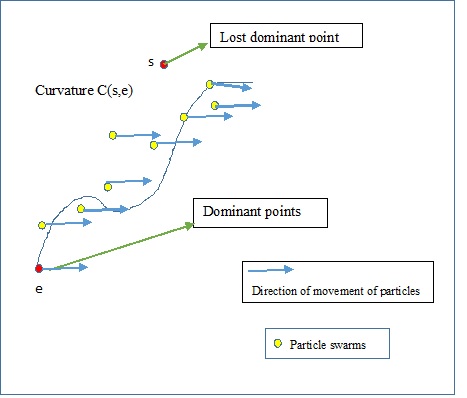}
	\caption{Lost dominant point which is shifted left is shown by RED color. Blue arrows show the direction of movement of the object. One of the swarm particles shown by yellow dot is lost;but other particles successfully track the curvature.}
\end{figure}

We can see that, in figure-(11)  KLT missed tracking of dominant point s. We also assume that some of the swarms may be lost during tracking because they are distributed over the path which originates from point s. This phenomenon is shown by yellow dots near to the lost dominant point.

In figure-(12) we show how re-initialization happens. The PSO algorithm tracks the curvature of the object which is approximated by a straight line between s and e. After few frames when we observe that s is not moving as its pixel position is not changing, we then consider that KLT has lost point s. When we detect that loss, we need another dominant point for continuation of our curvature tracking. But we do not compute another dominant point using formula-3 \cite{Ray92} as in real time tracking re-computation of lost dominant point is not a feasible solution. Instead of computing of another dominant point we assign a moving PSO particle which is nearest to the lost dominant point s. As we already keep tracking of all particles which follow the path joining between two dominant points,it is much more feasible and viable approach to follow. It does not require any computation and also we do not need to find where the closest particle will be.

\begin{figure}[h]
	\centering
	\label{fig:12}
	\includegraphics[width=0.6\textwidth]{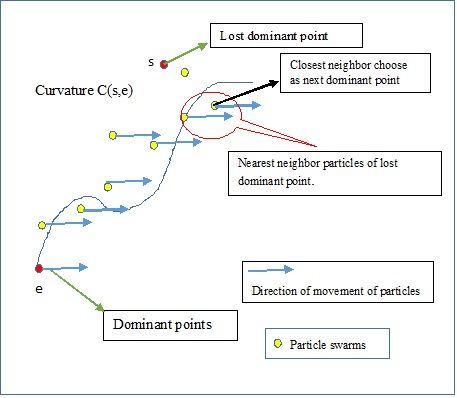}
	\caption{Nearest neighboring point selected as new dominant point. Other particles track as usual without considerable amount of delay.}
\end{figure}
In figure-(13) we have shown, after selection of a new dominant point, the entire curvature tracking is resumed.

By this approach, neither we have lost our tracking nor we have made any delay/break in tracking due to loss of dominant point. But the question is, weather the approach to replace lost dominant points by a new one rather then actual computation of dominant point is feasible? We need to remember we are not tracking exact contour of tracked object. Hence we do not need to follow exact curvature of the object body. Rather PSO Particles are tracking the approximated curvature of the object by simply a straight line joining two dominant points in our case. Though the newly selected point is not exact dominant point but it can easily solve our purpose to follow the object boundary. This heuristic approach to design a tracker is basically an attempt to extract the element of intelligence of a swarm.

\begin{figure}[h]
	\centering
	\label{fig:13}
	\includegraphics[width=0.6\textwidth]{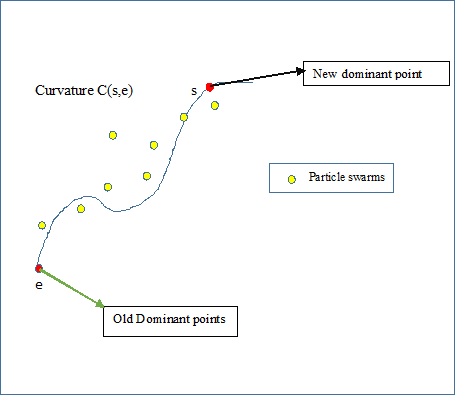}
	\caption{Newly selected point is marked as s and curvature tracking is resumed as earlier.}
\end{figure}
\subsection{Reinitialization of the particles of the swarms}
\label{Sec:312}
Reinitialization of particle(s) of individual swarm is sometime required as it is inherent in nature of PSO that few particles are too diverged from their desired position and even after several updation may not bring them towards their goal. In our case it is also possible that some particles are too far away from curvature boundary and after a finite number of iteration they still unable to converge. Then we need to reinitialize those particle. Reinitializing particles over entire image space certainly feasible but not a practical idea, because it again may diverge. So we have a better possibility to converge by assigning the position of the diverged particle on the current position of dominant point. Diagrammatically we can represent this phenomenon in figure-(14)

\begin{figure}[h]
	\centering
	\label{fig:14}
	\includegraphics[width=0.8\textwidth]{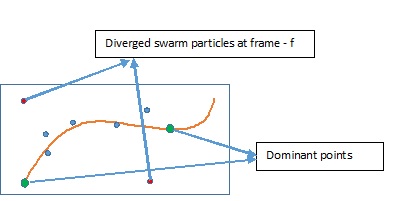}
	\caption{At frame number-f we detect two swarm particles marked as RED points are diverged.}
\end{figure}

Let us consider two swarm particles start diverging at frame number f and we detect this after few more frames are processed. Let say at frame (f + t), we find two particles are diverged. In frame (f+t+1) we took action about its repositioning. Lots of research works have been done about repositioning of diverged particle \cite{Roh11}, \cite{Ric04}, \cite{Dem12}. In \cite{Ric04}, Richards et.al use generators from centroidal Voronoi tessellations as the starting points for the swarm. In \cite{Dem12}, de Melo et.al consider the algorithm named Smart Sampling (SS) finds regions with high possibility of containing a global optimum. A meta-heuristic can be used to initialize inside each region to find that optimum. Smart Sampling(SS) and Differential Evaluation (DE) are combined to establish  SSDE algorithm to evaluate the approximate position of the diverged particles. So we can choose and apply any of these methods which works successfully. But in the present context, instead of doing this we consider another approach we think to be more effective in our case.

At frame number (f+t+1), we simply consider the X and Y direction of those diverged particles  and update their position according to the following formula.
\begin{equation}
p\_new_{ik}^x = dompts_i^x
\end{equation}
\begin{equation}
p\_new_{ik}^y = dompts_i^y
\end{equation}
where,
$p\_new_{ik}^x$ - is the latest X directional positional value of kth diverged particle from $i^{th}$ swarm.\\
$p\_new_{ik}^y$ - is the latest Y directional positional value of kth diverged particle from $i^{th}$ swarm.\\
$dompts_i^x$ - X directional value of the dominant point of the $i^{th}$ swarm.
$dompts_i^y$ - Y directional value of the dominant point of the $i^{th}$ swarm.

We need to keep in mind that, dompts is a set of dominant point which continuously updated frame by frame as per our algorithm. So whenever we mention dompts we always refer latest updated dominant points. Another point worth referring here is that swarm particles try to converge over the straight line joining two dominant points. So whenever any swarm particle is diverged from its desired location over a particular line segment, if we directly place the said diverged particle on any dominant points of the line segment, according to the above formula  then the question is which dominant point we should choose among two dominant points joining which we get the line segment?. Actually it does not produce any serious impact if we choose arbitrarily any dominant point among these two.

Basic intuition behind the above equations is that if some particles diverge, their main objective  is to reach as near as possible to the straight line joining two consecutive dominant points. So rather computing any complex mathematical function and performing extensive number of iteration we consider the most simple approach by placing the diverged particles positions directly on any of the two dominant points around which a swarm was already formed and from where particle(particles) were diverged.
Figure-(15) explains the newly updated position of the diverged particles.

\begin{figure}[H]
	\centering
	\label{fig:15}
	\includegraphics[width=0.8\textwidth]{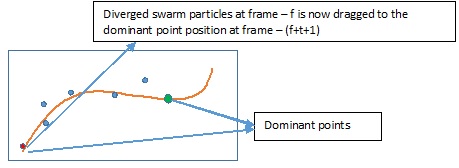}
	\caption{Convergence of the diverged swarm particle at frame (f+t+1).}
\end{figure}

Let us consider the following issue;\\
Thus random sampling of the line segment between two dominant points itself provides us the same effect as PSO ? \\The answer is many fold as stated below :\\
\begin{itemize}
\item We need to remember that, dominant points are basically those points which hold high amount of curvature information. Now for a moving objects, especially fast car or running human body etc, there are so many dominant points that if we consider most of them then entire process of tracking become cumbersome and clumsy. So we need to be careful during choosing dominant points. Now if we minimize the distance between two dominant points, then for a whole moving object we need huge number of dominant points. That is why we keep number of dominant points minimum and distance between two dominant points are not very close.
	
\item Now the approximate curvature between two dominant points is not exactly a line segment. It could be any curve. That is why random sampling is not giving any better results.
	
\item As the two dominant points are not so close to each other, the curvature between two dominant points is captured by the particles of the swarm; that is the main reason we need PSO.
	
\item From frame to frame, we may loose dominant points during tracking, then we don’t need to calculate dominant points again and again. The particles close to the dominant points become next dominant points. Thats why using PSO is beneficial rather than random sampling between two dominant points.

\end{itemize}

\section{Some further illustration on the proposed dual tracking algorithm.}
\label{sec:4}
\textbf{Step 1:} First we extract the first frame from input video and convert it into binary image.By trial and error we find a pixel point on the boundary of target object as shown in figure-(16). Here $P_{i}$($x_{i},y_{i}$) is the boundary point. Note that for simplicity of illustration we consider the front view of an object. In practice it can be any given aspect of an object to be tracked.
\begin{figure}[H]
	\centering
	\label{fig:16}
	\includegraphics[width=0.6\textwidth]{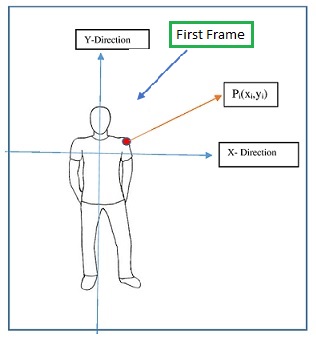}
	\caption{A human body where$P_{i}$($x_{i},y_{i}$) is mapped using trial and error method.}
\end{figure}

\textbf{Step 2:} Apply Freeman Chain code to find the breakpoints started with $P_{i}$ as shown in figure-(17).
\begin{figure}[h]
	\centering
	\label{fig:17}
	\includegraphics[width=0.6\textwidth]{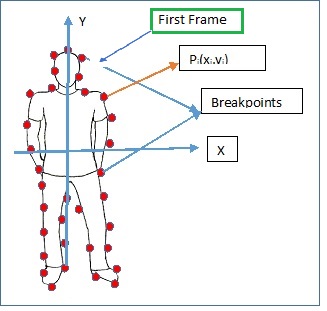}
	\caption{Boundary points are detected using freeman chain code(see equation-(1)).}
\end{figure}

\textbf{Step 3:} Find out dominant points using max cosine values(Ray,1992). Initially it eliminates all linear points and subsequently find out those points which have maximum cosine values.We denote those dominant point set as D.  =\{$D_{1},D_{2},D_{3}...D_{15}$\}. The resultant figure is shown in figure -(18).
\begin{figure}[h]
	\centering
	\label{fig:18}
	\includegraphics[width=0.6\textwidth]{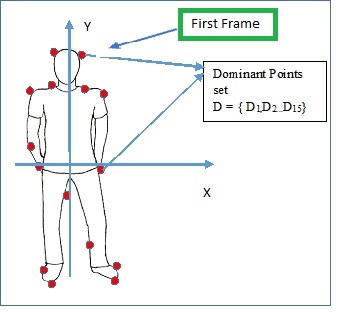}
	\caption{Selected Dominant points calculated from breakpoints are shown here (see equation (2) and (3))}
\end{figure}

\textbf{Step 4:} These dominant points as shown in figure-(18) by RED dots are tracked by KLT tracker from $frame_{1}$ to $frame_{n}$. In figure-(19) blue arrows show how KLT tracks dominant point independently from frame to frame.On $frame_{2}$ we distribute PSO particles over the image space randomly. The distribution of PSO particles shown in figure -(20). Again for simplicity of illustration we show front view of an object in Frame-2. In practice it can be any given aspect of an object as frame-2.

\begin{figure}[h]
	\centering
	\label{fig:19}
	\includegraphics[width=0.6\textwidth]{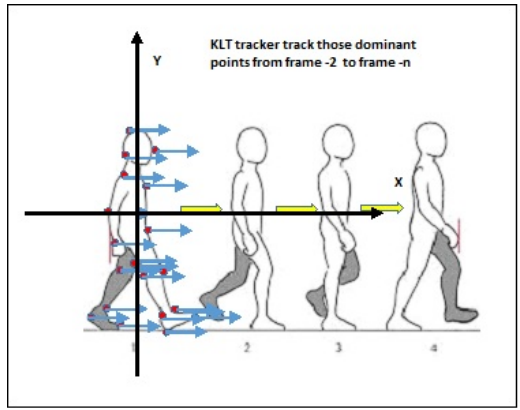}
	\caption{Tracking of dominant points as shown by Red Dots,by KLT tracker from one frame to another.}
\end{figure}

\begin{figure}[h]
	\centering
	\label{fig:20}
	\includegraphics[width=0.6\textwidth]{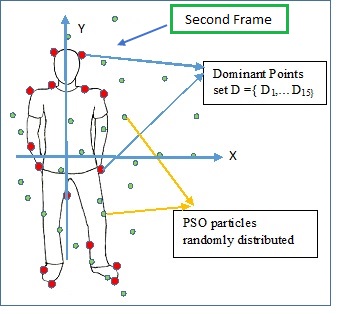}
	\caption{PSO particles distributed over the image space randomly }
\end{figure}

\textbf{Step 5:} On frame -2  as stated above and as shown in figure-(21) we draw the lines  joining two consecutive dominant points. All green dots are PSO particles and RED dots are dominant points. The straight line joining two consecutive  dominant points has been shown by black straight line and green PSO particles spread on those straight lines. The yellow arrows show the movement of PSO particles. The right hand object of figure-(21) is basically a polygonal approximation of the left hand object of figure -(21). The vertices's of the polygonal approximation(i.e the right hand object) represent the dominant points of the object at frame -2.
\begin{figure}[H]
	\centering
	\label{fig:21}
	\includegraphics[width=0.6\textwidth]{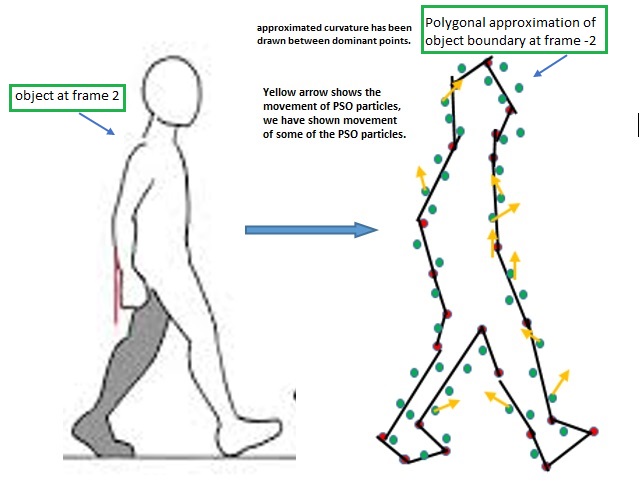}
	\caption{Polygonal approximation of the Object Boundary at frame 2.}
\end{figure}

\textbf{Step 6:} In figure-(22) we have shown that from frame -2  to frame -3 optical flow and swarm intelligence simultaneously performing the task of dual tracking. KLT(based on the concept of optical flow) tracks the dominant points of the object(i.e.the vertices's of the polygon)and PSO(based in the concept of Swarm Intelligence)tracks the boundary (approximated by the straight line of the polygon) of the object.\\
The green points are PSO particles and they are distributed over a straight line between two consecutive dominant points. The green points of PSO particles which are distributed over each small line segments of the dynamically generated polygon of the target object form a swarm. Thus around the polygonally approximated target object a multi-swarm scenario is generated and the approximated polygon of the target object is embedded over the annular ring (strip) of the multi-swarms (see figure (1)and(2)). Dominant points are marked as RED. Blue arrows show that the tracking of dominant points is performed by KLT.Yellow arrows show the tracking of the boundary(approximated by a straight line)of the curved object by PSO particles.

\begin{figure}[H]
	\centering
	\label{fig:22}
	\includegraphics[width=0.8\textwidth]{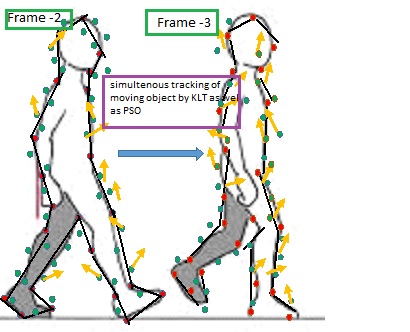}
	\caption{Simultaneous tracking of moving object by both KLT and PSO.}
\end{figure}

\begin{figure}[H]
	\centering
	\label{fig:23}
	\includegraphics[width=0.8\textwidth]{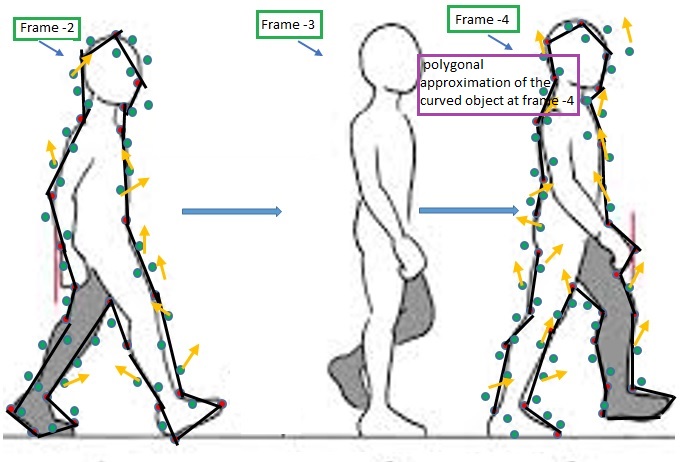}
	\caption{Object at frame-4 shows the polygonal approximation dynamically created and tracked by PSO particles.}
\end{figure}

\textbf{Step 8:} From frame 2 upto the last frame of the video sequence a bounding box, as shown in figure-(24), is designed based on position of PSO particles.
\begin{figure}[H]
	\centering
	\label{fig:24}
	\includegraphics[width=0.8\textwidth]{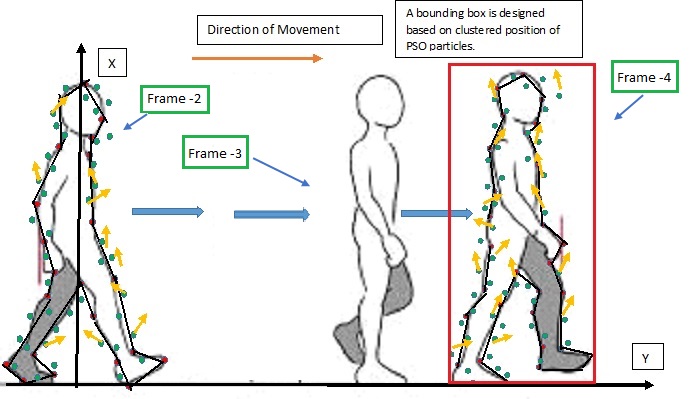}
	\caption{Bounding box is designed based on PSO particles position(see equation-(26)).}
\end{figure}

\textbf{Step 7:} Note that by process of dual tracking when the target object, which is polygonally approximated , reaches the 3rd frame of the video sequence the shape of the annular ring(strip) of the multi-swarm changes due to the change of the shape of the dynamically generated polygon for the movement of the non-rigid target object. The newly generated polygon of the target object is automatically embedded over the changed annular ring(strip) of the multi-swarm and the process of dual tracking proceed from frame-3 to last frame. Figure-(23) shows the polygon created by PSO particles at frame no.4. Thus, in addition to the tracking of the dominant points of the object, which are basically vertices's of the polygon dynamically created due to the change in shape of the nonrigid object(see the discussion on 3rd paragraph of section 2.1 and frame -4 of figure-23), we simultaneously track the boundary of the dynamically changed curved object which is approximated by straight line segments of the newly generated polygon(see frame-4 of figure-23) by PSO particles.

\section{Algorithmic summary and Complexity analysis}
\label{sec:5}
The Dual Tracking Algorithm(DTA) is represented in pseuducode as follows  -

\subsubsection{Pseudocode}
\label{sec:51}
See page-25 for the detail algorithm.
\begin{algorithm*}
	\caption{DualTrackingAlgorithm}
	\begin{algorithmic}[1]
		\Procedure{DTA(videoSequence\_with\_traget\_object)}{}
		\State $\text{Frames} \gets \text{CALL Algorithm \textit{\textbf{Frame\_Extraction(videoSequence\_with\_traget\_object)}}}$
		\Comment {\textit{[Get Frames form the input video; [see Appendix]]}}
		\State $\text{brpts} \gets \text{CALL Algorithm \textit{\textbf{BrPtCal(Frames)}}}$
		\Comment {\textit{[Calculate Breakpoints of target objects and store those points in “brpts” variable].[see Appendix]}}
		\State $\text{dompts} \gets \text{CALL Algorithm \textit{\textbf{DominantPt(brpts)}}}$
		\Comment {\textit{\textit{[Calculate dominant points from Breakpoints “brpts"]}.[see Appendix]}}
		\State $\text{nSwarm} \gets \text{number of swarms.}$
		\State $\text{ss} \gets \text{Number of particles per Swarm.}$
		\Comment {\textit{\textit{[Define number of swarms and number of particles in each swarm.]}}}
		\For {$ \text{particles} \gets 1 $ to ss$  $}
		\For {$ \text{pi} \gets 1 $ to ss$  $}
		\State $\text{Initialize particles pi velocity and position.}$
		\State $\text{Initialize plbest and pgbest.}$
		\State $\text{Compute  Procedure \textit{\textbf{FitnessComputePSO(pi)}}.}$
		\Comment {\textit{\textit{[Initialization of each particle’s position and velocity for the swarm. ]}}}
		\EndFor
		\EndFor 
		\For {$ \text{frame} \gets 1 $ to Frames$ $}
		\For {$ \text{dmpt} \gets 1 $ to dompts$ $}
		\State $\text{old\_dmpt} \gets \text{dompts(dmpt)} $
		\State $\text{dominantpts} \gets \text{CALL Algorithm \textit{\textbf{klt(dmpt)}}}$
		\Comment {\textit{[see Appendix]}}
		\If{dominantpts(dmpt)-Old\_dmpt(dmpt) = 0}
		\State $\text{NewDmpts} \gets \text{\textit{\textbf{DominantPointReInitialization(AcceptedParticles, dompts}}}$
		\Comment {\textit{[dominant point re-initialization.see Appendix]}}
		\EndIf 
		\EndFor
		\EndFor
		\For {$ \text{swarm} \gets 2 $ to nSwarm$ $}
		\For {$ \text{pi} \gets 1 $ to ss$ $}
		\State $\text{AcceptedParticles} \gets \text{\textit{\textbf{FitnessComputePSO(pi)}}}$
		\If {pi NOT in AcceptedParticles }
		\State $\text{Update velocity  of pi using Eq–(2)}$
		\State $\text{Update position  of pi using Eq–(1)}$
		\EndIf
		\EndFor
		\EndFor
		
		\State $\text{CALL Algorithm \textit{\textbf{BoundingBox(AcceptedParticles)}}}$
		\Comment {\textit{[Draw Bounding Box based on particles position available from "AcceptedParticles" vector]}}
		\Comment {\textit{[See Appendix]}}
		\EndProcedure
	\end{algorithmic}
\end{algorithm*}

\subsubsection{Complexity analysis}
\label{sec:52}
To calculate time complexity of Procedure \textit{\textbf{\\DualTrackingAlgorithm(DTA)()}} we need to compute complexity of all the sub algorithms it called and summing up all those complexity will give us approximated time complexity of this algorithm.

Complexity of Algorithm \textit{\textbf{Frame\_Extraction(\\Video\_input)}}$-$ reading a video file and extracting each frame and storing them in a separate file requires O(f). Where f is the number of frames.

Complexity of Algorithm \textit{\textbf{BrPtCal(Frames)}} - if the targeted object contains p number of pixels for the entire boundary, then freeman chain code at maximum will check 8 direction for boundary condition for each pixels. So at maximum the time required to find all the breakpoints for the object is – O(7*p), which we can consider as linear time O(q), where q approximately 7*p.

Complexity of Algorithm \textit{\textbf{DominantPt(brpts)}} – let’s consider number of breakpoints are – b, \\then no\_region = b/[5-10], lets that is $b_{1}$. Now computing k$-$cosine values required constant time. So final time required is O($b_{1}$ * b * k), where k is a constant time, $b_{1}$- no\_region, b – no of breakpoints.

Complexity of Algorithm \textit{\textbf{klt(dmpt)}}$-$ Assume that the number of warp parameters is n and the number of pixels in T is N. The total computational cost of each iteration of Lucas-Kanade algorithm is O($n^{2}$ * N + $n^{3}$), detail discussion explained in \textit{Lucas-Kanade 20 Years On: A Unifying Framework: Part 1}(Sundaram,2010).

Complexity of Algorithm \textit{\textbf{BoundingBox(\\AcceptedParticles)}}$–$ In order to get time complexity of this algorithm we need only to calculate time complexity of height and breadth procedure. According to algorithm it will be O(h) where is h is number of particle we need to check whether they lie on the object boundary range. Similarly for breadth it will be O(br) where br is b number of particle in breadth computation .so over all in \textit{\textbf{BoundingBox()}} algorithm time complexity will be O(b+h).

Complexity of Algorithm \textit{\textbf{DominantPointReInitialization(AcceptedParticles, dompts(dmpt))}} -\\ If total number of accepted particle is n and for sorting that vector containing x particle will at best take O(nlogn). After sorting we will took first particle as next new dominant point, so overall time complexity will ne O(nlogn).

Complexity of Algorithm \textit{\textbf{DominantPointReInitialization(AcceptedParticles, dompts(dmpt))}} -\\ If total number of accepted particle is n and for sorting that vector containing x particle will at best take O(nlogn). After sorting we will took first particle as next new dominant point, so overall time complexity will ne O(nlogn).

Final complexity of \textit{\textbf{DualTrackingAlgorithm(DTA)}}\\- Frame\_Extraction(),BrPtCal(), and DominantPt() will be called only one time so time required to compute will be -\\
=$\mathcal{O}(f)$ + $\mathcal{O}(q)$+ $\mathcal{O}($b\_{1}$* b * k)$\\
=O(f) + O(q) +O($b_{1}$*b) [ k is constant]\\
=O(f) + O(q) + O($b_{2}$) [ $b_{1}$ is very less than b]\\
=O(f+q) + O($b_{2}$) = O($b_{2}$)

Initialization of PSO particles will give us time O(n) where n is no. of particles. Now KLT will be called for every frame, so if there is f number of frame then this will give -  f * O($n_{2}$*N + $n_{3}$).and \textit{\textbf{DominantPointReInitialization()}} will called very few times so it is approximately O(nlogn),where n is the number of particles.And finially PSO will run for each frames required O(f), f is frame number.

Total time complexity = \textbf{[O($b_{2}$) + f* O($n_{2}$*N +$n_{3}$) + O(f)]} where b – no. of breakpoints, f – no of frames and n – number of pso particle, and N – no. of pixels in KLT tracking.

\section{Experimental Result and Analysis.}
\label{sec:6}
\subsection{Experimental Setting.}
\label{sec:61}
The proposed dual tracking approach for variable background and static background under different challenges as stated earlier, is tested by MatLab 2015a on a 64 bit PC with Intel i5 processor with 3 GHz speed. The image size of the frame 180 X 144. Static video is 20 sec duration whereas variable background is 13 sec duration.

\subsection{Experimental Dataset-1 (Wu et.all).\cite{Wu13}}
\label{sec:62}
All the experimental dataset has been taken from benchmark library created by Wu, \textit{Yi Wu, Jongwoo Lim and Ming-Hsuan Yang}\cite{Wu13} and \cite{Wu003} which is available on \url{http://cvlab.hanyang.ac.kr/tracker_benchmark/}.

\subsubsection{Tracking Results of the proposed method.}
\label{sec:621}
Form the experimental test data set we pick up 5 video streams which have static background and point of interest is moving high to moderate rate. From TB-50 sequence - \\
1.Girl[SV,OCC,IPR,OPR],\\
2.Walking2[SV,OCC,LR],\\
3.Walking[LR,IV],\\
4.FaceOcc1[OCC],\\
5.Boy[SV,FM,IPR,OPR]

and video stream from TB-100 sequence for dynamic background; \\
1.jogging(1)(2)[OCC,DEF,OPR],\\
2.Suv[OCC,IPR,OV],\\
3.Walking[OCC],\\
4.Skater2[SV,DEF,FM,IPR,OPR],\\
5.Football1[IPR,OPR].\\

Here,
\\
SV $-$ Scale Variation,\\
OCC $-$ Occlusion,\\
DEF $-$ Deformation,\\
IPR $-$ In-Plane –Rotation,\\
OPR $-$ Out-Plane-Rotation,\\
OV $-$ Out of View,\\
BC $-$ Background Clutter,\\
IV $-$ Illumination Variation,\\
LR $-$ Low Resolution 

are the attributes we consider.

Based on these video streams we demonstrate the tracking results of the proposed dual tracking algorithm. 

\subsubsection{Static background.}
\label{sec:622}
The first experiment is on static background. We consider - \\
1.Walking [LR,IV]\\
2.Girl[SV,OCC,IPR,OPR]\\
3.Walking2[SV,OCC,LR]\\
4.FaceOcc1[OCC]\\
5.Boy[SV,FM,IPR,OPR]

datasets in figure-(25) to figure-(34).
%\begin{multicols}{2}

\begin{figure*}[!ht]
	\centering
	\label{fig:25}
	\includegraphics[width=1.2\textwidth]{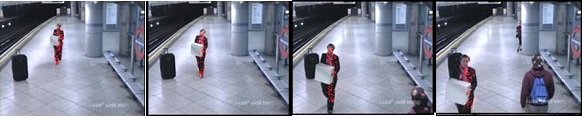}
	\caption{It shows a sequence of frames of a single person moving towards a camera where background is static. Proposed dual tracking algorithm successfully tracks the target object as it moves in Walking dataset. Green dots show the dominant points and RED dots show the swarm particles.}
\end{figure*}

\begin{figure*}[!ht]
	\centering
	\label{fig:26}
	\includegraphics[width=1.2\textwidth]{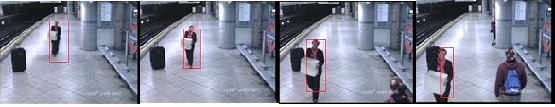}
	\caption{In Walking dataset, a bounding box based on PSO based algorithm is shown. For representational clarity the dominant points and swarm particles are not shown explicitly inside the bounding box.Comparison With other State-of-art algorithms shown here. RED box indicate DTA, YELLOW box ASLA \cite{Jia12},GREEN box CPF \cite{Per02}, BLUE box CT \cite{Zha12}.}
\end{figure*}

\begin{figure*}[!ht]
	\centering
	\label{fig:27}
	\includegraphics[width=1.2\textwidth]{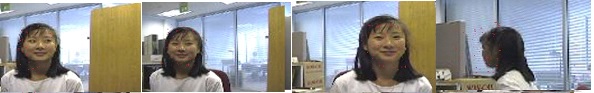}
	\caption{The Dual tracking approach tracks the movement of the face in GIRLS dataset. Green dots show the dominant points and RED dots show the swarm particles.}
\end{figure*}

\begin{figure*}[!ht]
	\centering
	\label{fig:28}
	\includegraphics[width=1.2\textwidth]{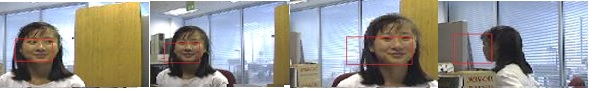}
	\caption{The Bounding Box is shown around the face of GIRLS dataset.For representational clarity the dominant points and swarm particles are not shown explicitly inside the bounding box.Comparison With other State-of-art algorithms shown here. RED box indicate DTA, YELLOW box ASLA \cite{Jia12},GREEN box CPF \cite{Per02}, BLUE box CT \cite{Zha12}.}
\end{figure*}

\begin{figure*}[!ht]
	\centering
	\label{fig:29}
	\includegraphics[width=1.2\textwidth]{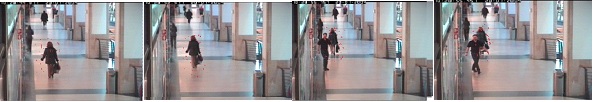}
	\caption{The Dual tracking approach tracks the movement of the target object in Walking2 dataset.Green dots show the dominant points and RED dots show the swarm particles.}
\end{figure*}

\begin{figure*}[!ht]
	\centering
	\label{fig:30}
	\includegraphics[width=1.2\textwidth]{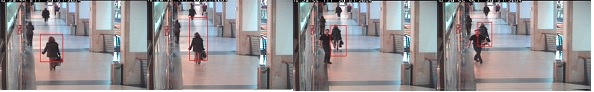}
	\caption{The Bounding Box is shown around the target object in Walking2 dataset.For representational clarity the dominant points and swarm particles are not shown explicitly inside the bounding box.Comparison With other State-of-art algorithms shown here. RED box indicate DTA, YELLOW box ASLA \cite{Jia12},GREEN box CPF \cite{Per02}, BLUE box CT \cite{Zha12}.}
\end{figure*}

\begin{figure*}[!ht]
	\centering
	\label{fig:31}
	\includegraphics[width=1.2\textwidth]{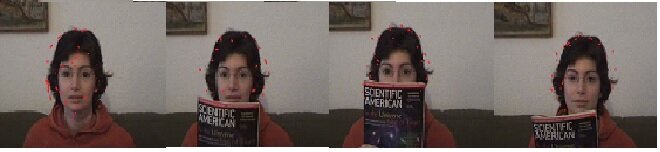}
	\caption{The Dual tracking approach tracks the movement of the target object in FaceOcc1 dataset.Green dots show the dominant points and RED dots show the swarm particles.}
\end{figure*}

\begin{figure*}[!ht]
	\centering
	\label{fig:32}
	\includegraphics[width=1.2\textwidth]{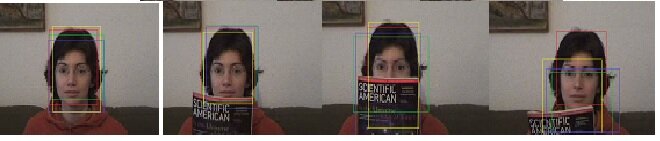}
	\caption{The Bounding Box is shown around the target object in FaceOcc1 dataset.For representational clarity the dominant points and swarm particles are not shown explicitly inside the bounding box.Comparison With other State-of-art algorithms shown here. RED box indicate DTA, YELLOW box ASLA \cite{Jia12},GREEN box CPF \cite{Per02}, BLUE box CT \cite{Zha12}}
\end{figure*}

\begin{figure*}[!ht]
	\centering
	\label{fig:33}
	\includegraphics[width=1.2\textwidth]{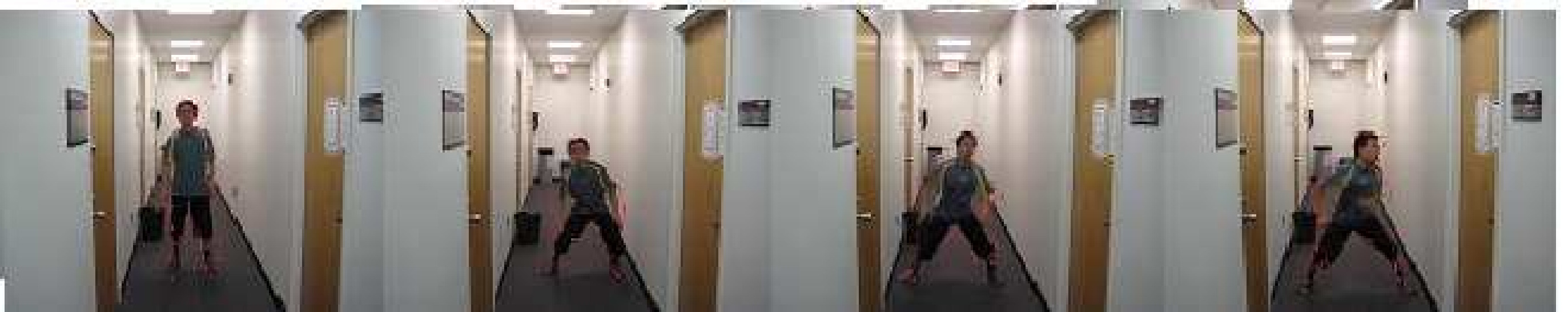}
	\caption{The Dual tracking approach tracks the movement of the target object in Boy dataset.Green dots show the dominant points and RED dots show the swarm particles.}
\end{figure*}

\begin{figure*}[!ht]
	\centering
	\label{fig:34}
	\includegraphics[width=1.2\textwidth]{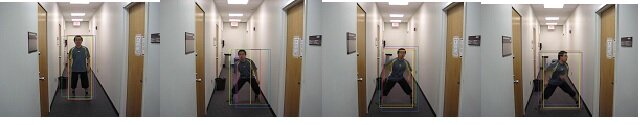}
	\caption{The Bounding Box is shown around the target object in Boy dataset.For representational clarity the dominant points and swarm particles are not shown explicitly inside the bounding box.Comparison With other State-of-art algorithms shown here. RED box indicate DTA, YELLOW box ASLA \cite{Jia12},GREEN box CPF \cite{Per02}, BLUE box CT \cite{Zha12}}
\end{figure*}

\subsubsection{Variable Background}
\label{sec:623}
Now we perform our experiment on a video where background is moving with object. Video is taken by a moving camera. Here we consider 3 video frames from TB-100 sequence namely -\\
1.jogging[1][2][OCC,DEF,OPR]\\
2.Suv[OCC,IPR,OV]\\
3.Walking[OCC,BC]\\
4.Skater2[SV,DEF,FM,IPR,OPR]\\
5.Football1[IPR,OPR,BC].

Both tracking results and Bounding Box representations are shown below from figure-(35) to (44).

\begin{figure*}[!ht]
	\centering
	\label{fig:35}
	\includegraphics[width=1.2\textwidth]{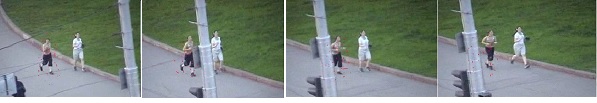}
	\caption{ The tracking result obtain by the dual tracking approach for jogging (1)(2) dataset.Green dots show the dominant points and RED dots show the swarm particles.}
\end{figure*}

\begin{figure*}[!ht]
	\centering
	\label{fig:36}
	\includegraphics[width=1.2\textwidth]{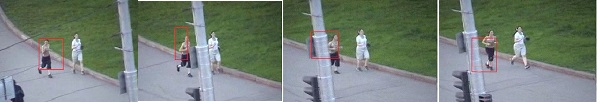}
	\caption{The bounding box representation is shown for jogging (1)(2) dataset. For representational clarity the dominant points and swarm particles are not shown explicitly inside the bounding box.Comparison With other State-of-art algorithms shown here. RED box indicate DTA, YELLOW box ASLA \cite{Jia12},GREEN box CPF \cite{Per02}, BLUE box CT \cite{Zha12}}
\end{figure*}

\begin{figure*}[!ht]
	\centering
	\label{fig:37}
	\includegraphics[width=1.2\textwidth]{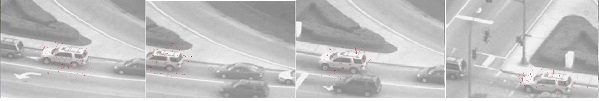}
	\caption{The tracking result obtained by the dual tracking approach for SUV dataset.Green dots show the dominant points and RED dots show the swarm particles.}
\end{figure*}

\begin{figure*}[!ht]
	\centering
	\label{fig:38}
	\includegraphics[width=1.2\textwidth]{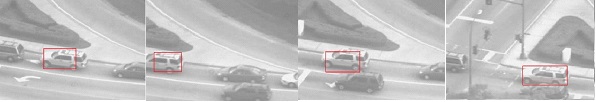}
	\caption{The bounding box representation is shown for SUV dataset.For representational clarity the dominant points and swarm particles are not shown explicitly inside the bounding box.Comparison With other State-of-art algorithms shown here. RED box indicate DTA, YELLOW box ASLA \cite{Jia12},GREEN box CPF \cite{Per02}, BLUE box CT \cite{Zha12}}
\end{figure*}

\begin{figure*}[!ht]
	\centering
	\label{fig:39}
	\includegraphics[width=1.2\textwidth]{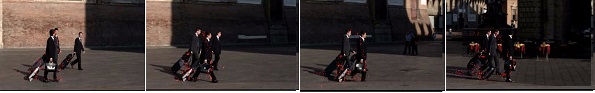}
	\caption{The tracking result is obtained by the dual tracking approach for Walking dataset. Green dots show the dominant points and RED dots show the swarm particles.}
\end{figure*}

\begin{figure*}[!ht]
	\centering
	\label{fig:40}
	\includegraphics[width=1.2\textwidth]{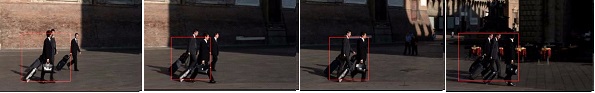}
	\caption{The bounding box representation is shown for Walking dataset. For representational clarity the dominant points and swarm particles are not shown explicitly inside the bounding box.Comparison With other State-of-art algorithms shown here. RED box indicate DTA, YELLOW box ASLA \cite{Jia12},GREEN box CPF \cite{Per02}, BLUE box CT \cite{Zha12}}
\end{figure*}

\begin{figure*}[!ht]
	\centering
	\label{fig:41}
	\includegraphics[width=1.2\textwidth]{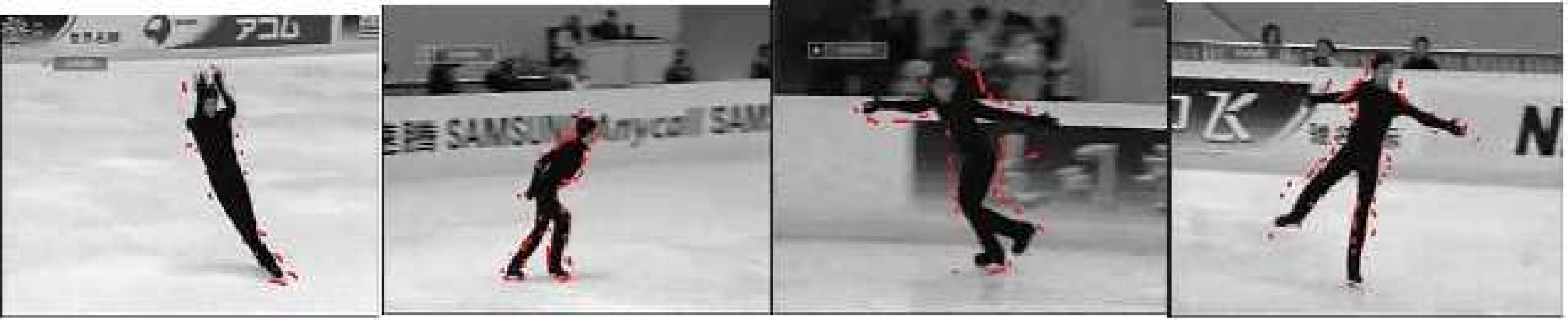}
	\caption{The tracking result is obtained by the dual tracking approach for Skater2 dataset. Green dots show the dominant points and RED dots show the swarm particles.}
\end{figure*}

\begin{figure*}[!ht]
	\centering
	\label{fig:42}
	\includegraphics[width=1.2\textwidth]{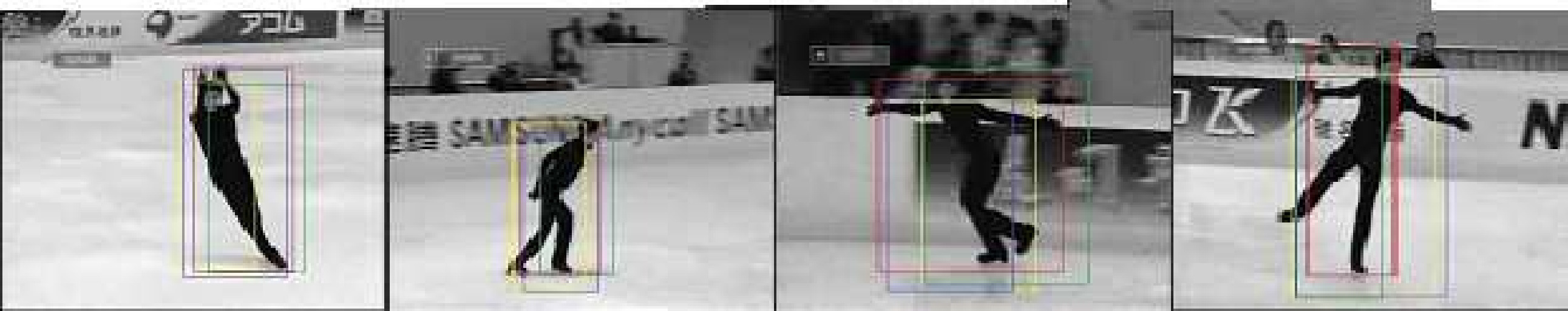}
	\caption{The bounding box representation is shown for Skater2 dataset. For representational clarity the dominant points and swarm particles are not shown explicitly inside the bounding box.Comparison With other State-of-art algorithms shown here. RED box indicate DTA, YELLOW box ASLA \cite{Jia12},GREEN box CPF \cite{Per02}, BLUE box CT \cite{Zha12}}
\end{figure*}

\begin{figure*}[!ht]
	\centering
	\label{fig:43}
	\includegraphics[width=1.2\textwidth]{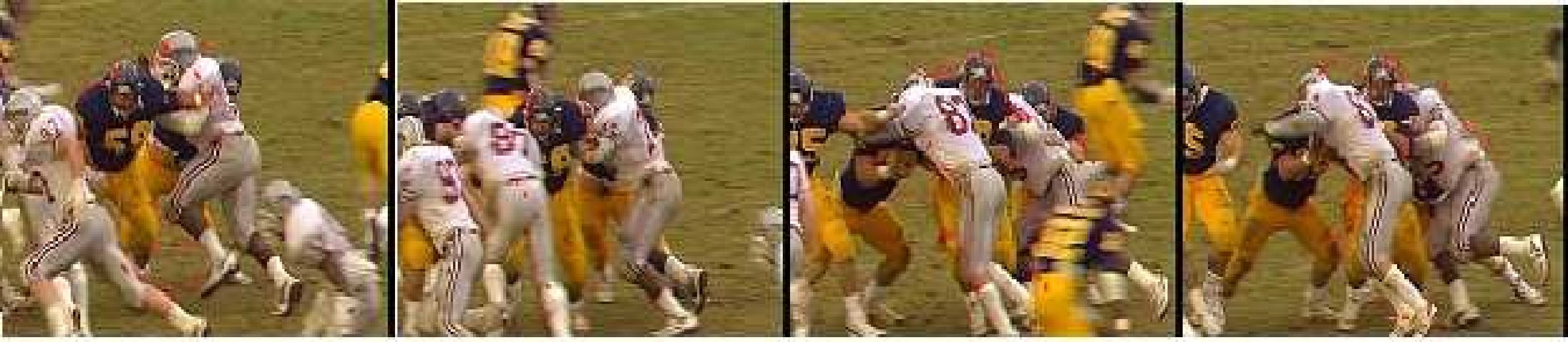}
	\caption{The tracking result is obtained by the dual tracking approach for Football1 dataset.Green dots show the dominant points and RED dots show the swarm particles.}
\end{figure*}

\begin{figure*}[!ht]
	\centering
	\label{fig:44}
	\includegraphics[width=1.2\textwidth]{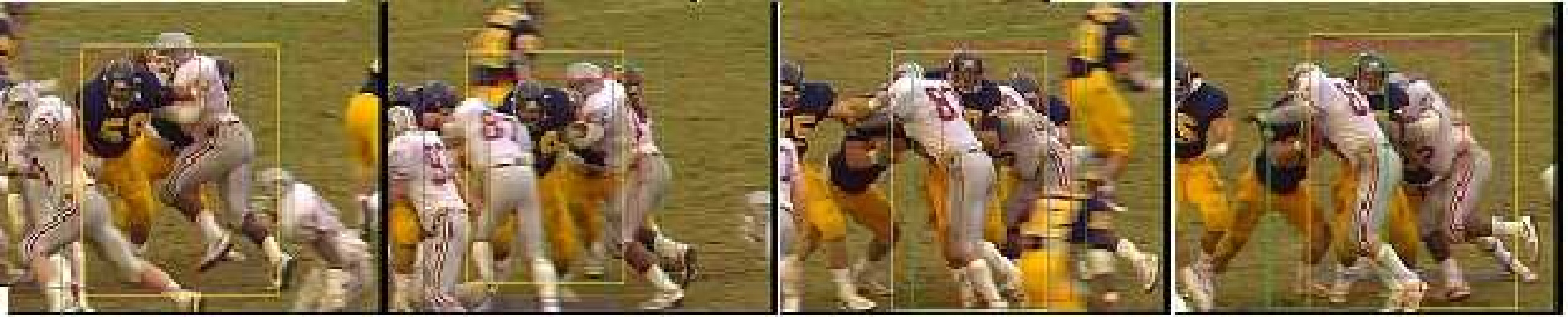}
	\caption{The bounding box representation is shown for Football1 dataset. For representational clarity the dominant points and swarm particles are not shown explicitly inside the bounding box.Comparison With other State-of-art algorithms shown here. RED box indicate DTA, YELLOW box ASLA \cite{Jia12},GREEN box CPF \cite{Per02}, BLUE box CT \cite{Zha12}}
\end{figure*}

We have tested the overall performance of our proposed Dual Tracking Algorithm(DTA) not only for the  above data sets but also for other datasets such as;\\
1.Dog[SV,DEF,OPR]\\
2.Football[OCC,IPR,OPR,BC]\\
3.Human2[IV,SB,OPR]\\
4.Human3[SV,OCC,DEF]\\
5.Girl[SV,OCC,IPR,OPR]\\
6.Singer1[IV,SV,OCC, OPR]\\
7.Skater2[SV,DEF,FM, IPR, OPR]\\
8.Women[IV, SV, OCC, DEF].

Tracking results have been shown in figure-(45).

\begin{figure*}[!ht]
	\centering
	\label{fig:45}
	\includegraphics[width=1.2\textwidth]{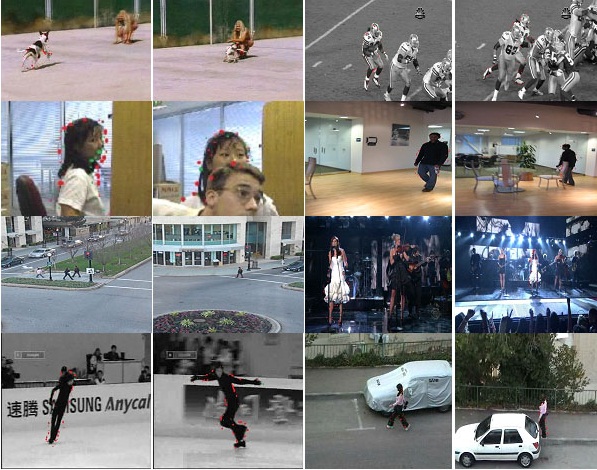}
	\caption{The tracking result obtained by the dual tracking approach for above mentioned datasets are shown.Green dots show the dominant points and RED dots show the swarm particles.}
	\end{figure*}

\subsubsection{Analysis and Evaluation}
\label{sec:624}
We evaluate the proposed dual tracking algorithm(DTA) using three parameters: True Detection(TD), False Detection(FD), Missed Detection(MD). We consider the parameter Frames per Seconds(FPS) to denote the number of frames per second. There is substantial amount of impacts in tracking due to high speed (high FPS) video sequence.
TD is evaluated as the percentage of frames that successfully detect object and track in each video sequence. Following is the mathematical formulation of TD :
\begin{equation}
TD=\frac{n_{td}}{N}\times100.
\end{equation}
where N = Total number of frames in a video sequence and $n_{td}$ = number of frames that qualify as Truly Detected objects. Successful object detection and tracking can be computed as per the following rule - we consider a frame is successful in marking  with the target object if our proposed bounding box around a detected and tracked object overlaps with the bounding box of the given ground truth. Mathematically, $|Ct_{to}-Ct_{gt}|\leq B_t/B_g$, where Ct's are the centroids and B's are the bounding boxes, $Ct_{tg}$ centroid of the ground truth,  $Ct_{to}$ centroid of the target object. In this ratio $B_t/B_g$, $B_t$ represents area of the proposed  bounding box of the target object and $B_g$ represents area of the bounding box of the ground truth. If the detected object position of our algorithm does not match with the position indicated by ground truth value or both of them do not overlap i.e. $|Ct_tg-Ct_gt|>B_t/B_g$; then the detection is false and is represented as: 

\begin{equation}
FD=\frac{n_{fd}}{n_{td}+n_{fd}}\times100.
\end{equation} 
If the algorithm is unable to detect target object in a frame, but ground truth value exists; then the situation is considered as Missed Detection and is represented as: 
\begin{equation}
MD=\frac{n_{md}}{n_{td}+n_{md}}\times100
.
\end{equation}

We have compared the proposed Dual Tracking Algorithm(DTA) with other state of the art algorithms: Visual tracking via adaptive structural local sparse appearance model(ASLA) \cite{Jia12}, Beyond semi-supervised tracking: Tracking should be as simple as detection, but not simpler than recognition(BSBT) \cite{Sta09}, Color-based probabilistic tracking(CPF) \cite{Per02}, Exploiting the circulant structure of tracking-by-detection with kernels(CSK) \cite{Hen12}, Real-time compressive tracking(CT) \cite{Zha12}.

Table-1 shows the comparative result of execution time with various tracking algorithm stated above based on FPS with respect to all six attributes stated above in dataset-1 \cite{Wu13} and \cite{Wu003}. In all cases We achieve superior results in comparison with other algorithms.

\begin{table*}[ht]
	%	\begin{threeparttable}
	\setlength\tabcolsep{12.3pt}
	\centering
	\begin{tabular}{ ||c|c|c|c|c|c|c|c|| }
		\hline
		\textbf{Attributes} & \textbf{ASLA}\textsuperscript{a} & \textbf{BSBT}\textsuperscript{b} & \textbf{CPF}\textsuperscript{c}  & \textbf{CSK}\textsuperscript{d} & \textbf{CT}\textsuperscript{e}  & \textbf{DTA}  \\
		\hline\hline
		\textbf{OCC} & 47 & 56 & 34 & 45 & \color{blue}67 & \color{red}78\\
		\textbf{SV}  & 57 & 45 & 32 & 45 & \color{blue}61 & \color{red}67\\
		\textbf{DEF} & 49 & 45 & \color{blue}67 & 33 & 45 & \color{red}71\\
		\textbf{OPR} & 62 & \color{red}72 & 56 & 55 & 65 & \color{blue}70\\
		\textbf{IPR} & 58 & 46 & 66 & \color{blue}71 & 56 & \color{red}74\\
		\textbf{BC}  & 59 & \color{blue}78 & 34 & \color{blue}78 & 56 & \color{red}87\\
		\hline
		
	\end{tabular}
	\begin{tablenotes}
		\item	\textsuperscript{a}(Jia et al.,2012),\textsuperscript{b}(Stalder,2009),\textsuperscript{c}(Perez et al.,2012),\textsuperscript{d}(henriques et al.,2012),\textsuperscript{e}(Zhang et al.,2012)
	\end{tablenotes}
	%\tabnote{\textsuperscript{a}(Jia et al.,2012),\textsuperscript{b}(Stalder,2009),\textsuperscript{c}(Perez et al.,2012),\textsuperscript{d}(henriques et al.,2012),\textsuperscript{e}(Zhang et al.,2012)}
	\caption{Attribute wise Execution time based on Frames per Second (FPS) on the benchmark datasets \cite{Wu003} and \cite{Wu13}.}
	\begin{tablenotes}
		\item {\color{red}red}: rank1, {\color{blue}blue}: rank2    
		
	\end{tablenotes}
	\label{table:1}
	%	\end{threeparttable}
\end{table*}

\begin{table*}[ht]
	\renewcommand{\arraystretch}{1.5}
	\label{table_er}
	\begin{threeparttable}
		\fontsize{7}{5}\selectfont
		\centering
		\setlength\tabcolsep{1.4pt}
		\begin{tabular}{||c||cccccc|cccccc|cccccc||} 
			\hline
			\multirow{2}{*}{\textbf{Attribute}} &
			\multicolumn{6}{c}{\textbf{Mean TD(\%)}} &
			\multicolumn{6}{c}{\textbf{Mean FD(\%)}} &
			\multicolumn{6}{c}{\textbf{Mean MD(\%)}} \\
			
			& \textbf{ASLA} & \textbf{BSBT}  & \textbf{CPF}  & \textbf{CSK} & \textbf{CT} & \textbf{DTA} & \textbf{ASLA}  & \textbf{BSBT}  & \textbf{CPF}  & \textbf{CSK} & \textbf{CT} & \textbf{DTA} & \textbf{ASLA}  & \textbf{BSBT}  & \textbf{CPF}  & \textbf{CSK} & \textbf{CT} & \textbf{DTA}  \\
			\hline\hline
			{\textbf{OCC}} & 77 & 73.77 & 72.36 & 72.9 & {\color{blue}77.5} & {\color{red}80.1} & 4.72 & 4.2 & 4.81 & 4.12 & {\color{blue}3.92} & {\color{red}3.63} & {\color{blue}2.0} & 3.41 & 3.67 & 3.12 & 3.41 & {\color{red}1.97} \\
			\hline
			\textbf{SV} & {\color{blue}85.23} & 80 & 78.1 & 71.1 & 79.23 & {\color{red}88.19} & 4.20 & {\color{blue}2.9} & 2.2 & 2.7 & 2.9 & {\color{red}2.0} & 3.17 &  {\color{red}1.2} & {\color{blue}1.43} & 2.8 & 2.91 & {\color{red}1.2} \\
			\hline 
			\textbf{DEF} & {\color{blue}70.63} & 67.21 & 67.92 & 60.51 & 69.86 & {\color{red}75.13} & 4.25 & 4.21 &  {\color{red}3.16} & {\color{blue}3.84} & 4.3 &{\color{blue} 3.27} & 3.17 & {\color{blue}2.92} & 2.86 & 3.62 & 3.19 & {\color{red}2.57} \\
			\hline 
			\textbf{OPR} & 60.45 & {\color{blue}68}  & 57.20 & 61.4 & 62.4 & {\color{red}69.23} & 5.67 & 5.91 & {\color{blue}5.57} & 5.9 & 5.7 & {\color{red}5.28} & 3.13 & 2.92 & 3.96 & {\color{blue}2.7} & 3.7 & {\color{red}2.4} \\
			\hline
			\textbf{IPR} & 60.24 & {\color{blue}61}  & 52.1 & 59.1 & 59.4 & {\color{red}65.23} & 5.7 & 5.81 & {\color{blue}5.77} & 6.3 & 6.7 & {\color{red}5.21} & 3.19 & 3.32 & 3.95 & {\color{blue}3.12} & 3.17 & {\color{red}3.0} \\
			\hline
			\textbf{BC} & 51.45 & {\color{blue}68}  & 57.1 & 60.1 & 62.4 & {\color{red}69.73} & 5.93 & 5.81 & {\color{blue}5.77} & 5.99 & 5.87 & {\color{red}5.28} & 3.13 & 3.92 & 3.83 & {\color{blue}2.6} & 3.9 & {\color{red}2.1} \\
			\hline
		\end{tabular}
		\caption{Attribute-wise Experimental Results on Benchmark Datasets \cite{Wu13}}
		\begin{tablenotes}
			\item {\color{red}red}: rank1, {\color{blue}blue}: rank2      
		\end{tablenotes}
	\end{threeparttable}
\end{table*}

\begin{table*}[ht]
	%	\begin{threeparttable}
	\setlength\tabcolsep{12.3pt}
	\centering
	\begin{tabular}{ ||c|c|c|c|c|c|c|c|| }
		\hline
		\textbf{Attributes} & \textbf{ASLA}\textsuperscript{a} & \textbf{BSBT}\textsuperscript{b} & \textbf{CPF}\textsuperscript{c}  & \textbf{CSK}\textsuperscript{d} & \textbf{CT}\textsuperscript{e}  & \textbf{DTA}  \\
		\hline\hline
		\textbf{SV} & 67 & \color{blue}78 & 54 & 35 & 77 & \color{red}91\\
		\textbf{MB}  & 47 & 45 & 62 & 70 & \color{blue}69 & \color{red}77\\
		\textbf{OCC} & \color{blue}79 & 66 & 77 & 53 & \color{red}95 & 71\\
		\textbf{MI} & 65 & 70 & 51 & \color{red}85 & 65 & \color{blue}71\\
		\textbf{OV} & \color{blue}68 & 54 & 46 & 51 & 56 & \color{red}69\\
		%\textbf{BC}  & 59 & \color{blue}78 & 34 & \color{blue}78 & 56 & \color{red}87\\
		\hline
		
	\end{tabular}
	\begin{tablenotes}
		\item	\textsuperscript{a}(Jia et al.,2012),\textsuperscript{b}(Stalder,2009),\textsuperscript{c}(Perez et al.,2012),\textsuperscript{d}(henriques et al.,2012),\textsuperscript{e}(Zhang et al.,2012)
	\end{tablenotes}
	%\tabnote{\textsuperscript{a}(Jia et al.,2012),\textsuperscript{b}(Stalder,2009),\textsuperscript{c}(Perez et al.,2012),\textsuperscript{d}(henriques et al.,2012),\textsuperscript{e}(Zhang et al.,2012)}
	\caption{Attribute wise Execution time based on Frames per Second (FPS) on the benchmark datasets-2(TLP dataset)\cite{Mou17}.}
	\begin{tablenotes}
		\item {\color{red}red}: rank1, {\color{blue}blue}: rank2    
		
	\end{tablenotes}
	\label{table:3}
	%	\end{threeparttable}
\end{table*}

For each of the above mentioned tracking algorithm , based on six attribute , True Detection(TD), False Detection(FD), Missed Detection(MD) are evaluated and presented in Table-2.

\subsection{Experimental dataset-2(TLP dataset)\cite{Mou17}}
Recently, Moudgil et.al developed a new benchmark dataset which contain long duration video sequence which they name as \textit{"Track Long and Prosper"} \textbf{(TLP)}. This dataset contain 50 real world videos which is approximately 400 minutes nearly 676K frames. This dataset is important because  most tracking algorithms work best in short sequences but drastically fail on long challenging video sequence. We perform experiment with our proposed dual tracking approach on 6 different dataset from this benchmark suite. This benchmark suite is available here: \textit{https://amoudgl.github.io/tlp/}

From the TLP dataset we pick up 6 video streams which have 5 attributes: Scale Variation(SV), Motion Blur(MB),Occlusion(OCC), Multiple Instances(MI),Out of view(OV). This 5 attributes are very challenging attributes as: OV indicates a situation where target fully out of the viewing window momentarily, similarly, MI indicates more than one objects with similar appearance as the target exist in the sequence and interact with it. This six video sequence with their attributes distributions are :\\
1.Lion(SV, MB, OCC, MI)\\
2.Badminton1(MB,OV,OCC,MI)\\
3.Boat(SV,OV,MB)\\
4.Carchase1(SV,OCC,MI,OV)\\
5.Helicopter(SV)\\
6.Jet5(SV,MB,OCC,OV).\\
This 6 datasets contains other attributes as well. We mention those attributes which we consider for comparison with the proposed dual tracking algorithm.
Figure-(46) shows tracking results on 6 challenging video streams consecutively.

Table-3 shows the comparative results of execution time (based on Frames per Second(FPS))with various tracking algorithms with respect to all Five attributes stated above in dataset-2(TLP dataset)\cite{Mou17}. In all cases We achieve superior results in comparison with other algorithms.

\begin{figure*}[!ht]
	\centering
	\label{fig:46}
	\includegraphics[width=0.9\textwidth]{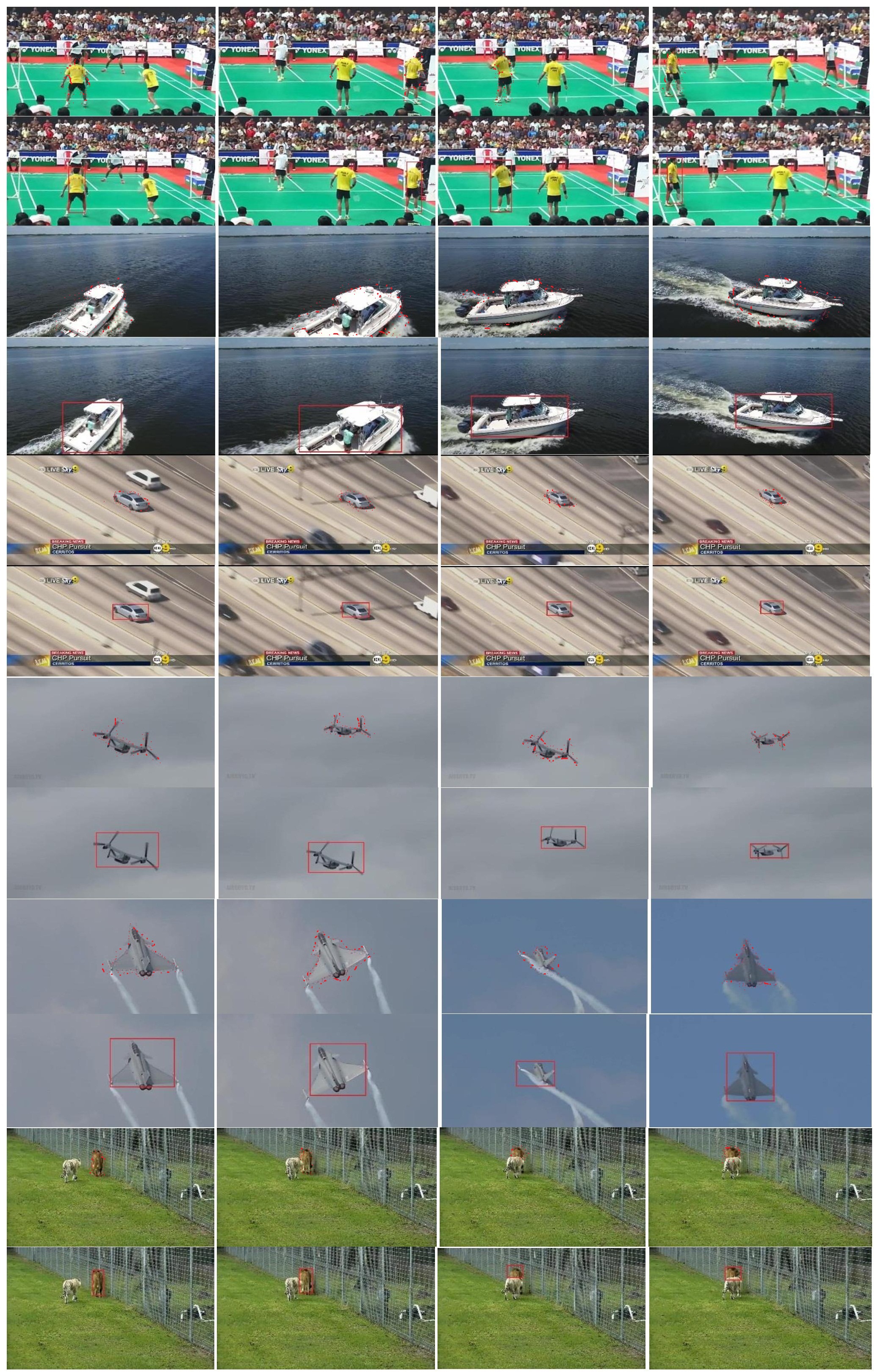}
	\caption{The tracking result obtain by the proposed Dual Tracking Algorithm(DTA) for above mentioned datasets.Green dot shows the dominant points and RED dots shows the swarm particles and Bounding box also present in every dataset.Comparison With other State-of-art algorithms shown here. RED box indicate DTA, YELLOW box ASLA \cite{Jia12},GREEN box CPF \cite{Per02}, BLUE box CT \cite{Zha12}}
\end{figure*}

\subsubsection{Analysis and Evaluation Methodology}
\label{631}
We further compare the proposed Dual Tracking Algorithm(DTA) in terms of three evaluation method: Precision plot, Success plot and Longest Subsequence Measure(LSM) \cite{Mou17}.

\textbf{Precision plot} - It is the most common and widely used method in object tracking \cite{Wu13} \cite{Gra08}. It shows the percentages of frames whose calculated pixel position(location) of the image is within the given threshold distance of the ground truth value. We use threshold distance value as 20 \cite{Bab11}.

\textbf{Success plot} - Another evaluation metric is success plot \cite{Wu13}. It provides the result of computing Intersection over Union(IoU) between computed and ground-truth bounding box position and also computes the number of successful frames whose IoU values is larger than given threshold values. If computed bounding box position of the target object  is - $BT_c$ and given groundtruth bounding box position value is - $GT_c$, then the overlap score \cite{Eve10} is - 
\begin{equation}
OS = \frac{|BT_c \cap GT_c|}{|BT_c \cup GT_c|} 
\end{equation}
where $\cap$ and $\cup$ represent intersection and union operators respectively and $| . |$ represents number of pixels in that region. We take Average Overlap Score(AOS) as the performance metric. AOS value decides weather a frame is successfully tracked or not.

\textbf{LSM plot} - It shows \cite{Mou17} which tracking algorithm successfully tracks the length of longest tracked subsequence per sequence. If F percentage of frames in a long video sequence is successfully tracked then we call it Longest Subsequence(LS), where F is an appropriate large value.

We pick up 5 state-of-art algorithm form the TLP dataset \cite{Mou17} : Learning Multi-Domain Convolutional Neural Networks for Visual Tracking(MDNet) \cite{Nam16}, Fully-convolutional siamese networks for object tracking\\(SiamFC) \cite{Ber16}, CREST: Convolutional Residual Learning for Visual Tracking(CREST) \cite{Son17},Action-Decision Networks for Visual Tracking with Deep Reinforcement Learning (ADNet) \cite{Yun17},MEEM: Robust Tracking via Multiple Experts using Entropy Minimization(MEEM) \cite{Zha14} We perform the task of comparison with the proposed Dual Tracking Algorithm(DTA) in terms of Success plot , Precision plot and LSM plot. The proposed Dual Tracking Algorithm(DTA) achieves superior result in all three categories. In figure-(47), we show the success plot and the precision plot and in figure-(48) we  show the LSM plot.

\begin{figure*}[!ht]
	\centering
	\label{fig:47}
	\includegraphics[width=1.3\textwidth]{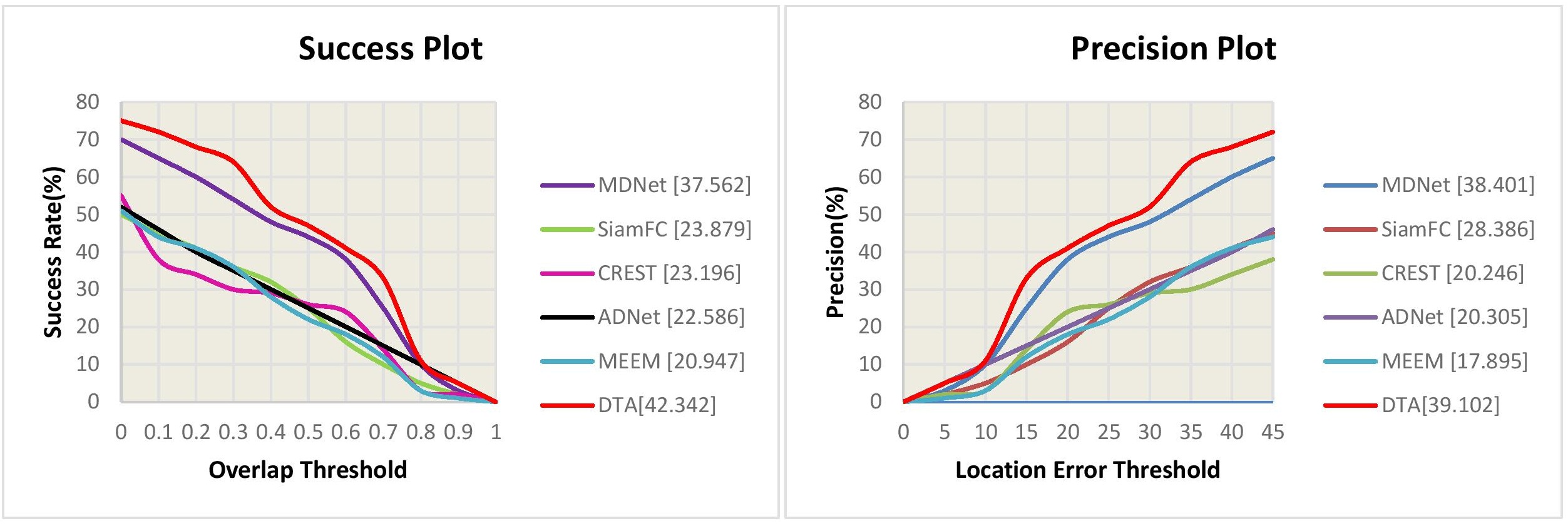}
	\caption{precision and Success plot evaluated on TLP dataset with Five other state-of-art algorithms.}
\end{figure*}

\begin{figure}[h]
	\centering
	\label{fig:48}
	\includegraphics[width=0.8\textwidth]{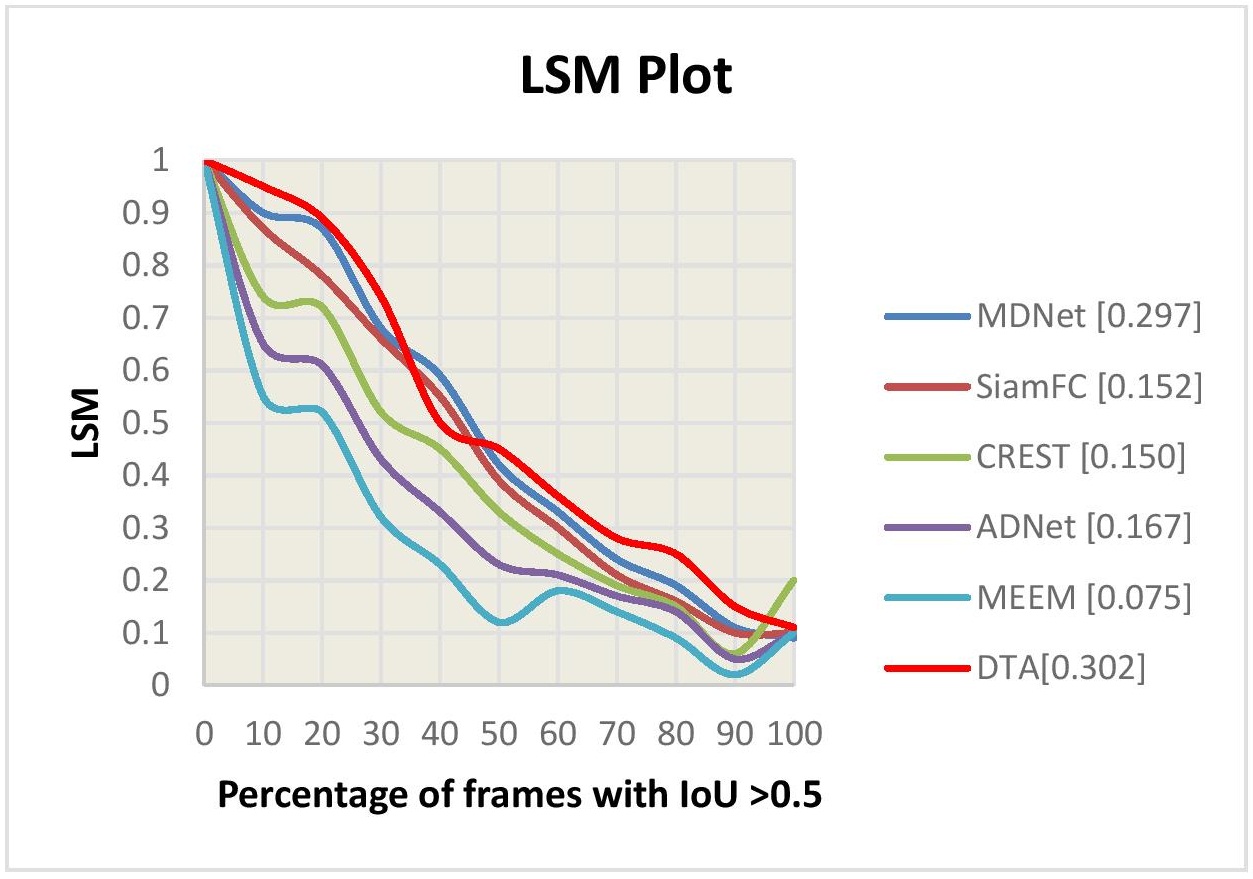}
	\caption{Longest Subsequence Measure(LSM) plots evaluated on TLP dataset with Five other state-of-art algorithms.}
\end{figure}

\subsection{Experimental dataset-3(Performance analysis with other PSO algorithms)}
\label{64}
We extend our experiment with other state-of-the-art Particle Swarm Optimization algorithms. We choose several published PSO algorithms  and compare their tracking performance with the proposed dual tracking algorithm(DTA). We consider the \textit{Object Tracking Evaluation 2012 from The KITTI Vision Benchmark Suite}\cite{Gei12}. Web link is as follows -- \url{http://www.cvlibs.net/datasets/kitti/eval\_tracking.php}.

Table-4 shows the comparative results of execution time(based on Frames per Second(FPS)) with various tracking algorithms with respect to all Five attributes stated above in dataset-3(KITTI Vision Benchmark Suite)\cite{Gei12}. In all cases We achieve superior results in comparison with other algorithms.

\begin{table*}[ht]
	%	\begin{threeparttable}
	\setlength\tabcolsep{20.3pt}
	\centering
	\begin{tabular}{ ||c|c|c|c|c|c|c|c|| }
		\hline
		\textbf{Attributes} & \textbf{A}\textsuperscript{a} & \textbf{B}\textsuperscript{b} & \textbf{C}\textsuperscript{c} & \textbf{D}\textsuperscript{d} & \textbf{E}\textsuperscript{e} & \textbf{DTA} \\
		\hline\hline
		\textbf{SV} & \color{blue}77 & 71 & 42 & 61 & 67 & \color{red}80\\
		\textbf{MB}  & 77 & 78 & 61 & \color{red}87 & 63 & \color{blue}83\\
		\textbf{OCC} & \color{blue}76 & 63 & 47 & 73 & \color{red}90 & 73\\
		\textbf{MI} & 75 & 72 & 61 & \color{blue}77 & 73 & \color{red}91\\
		\textbf{OV} & \color{blue}68 & 55 & 42 & 61 & 60 & \color{red}71\\
		%\textbf{BC}  & 59 & \color{blue}78 & 34 & \color{blue}78 & 56 & \color{red}87\\
		\hline
		
	\end{tabular}
	\begin{tablenotes}
		\item{\textsuperscript{a}(Hsu et al.,2012),\textsuperscript{b}(Boguslaw,2014),\textsuperscript{c}(Xuan et al.,2013),\textsuperscript{d}(Zhang et al.,2014),\textsuperscript{e}(Xia et al.,2017)}
	\end{tablenotes}
	%\tabnote{\textsuperscript{a}(Jia et al.,2012),\textsuperscript{b}(Stalder,2009),\textsuperscript{c}(Perez et al.,2012),\textsuperscript{d}(henriques et al.,2012),\textsuperscript{e}(Zhang et al.,2012)}
	\caption{Attribute wise Execution time based on Frames per Second (FPS) with other PSO algorithms on the benchmark datasets-3 \cite{Gei12}.}
	\begin{tablenotes}
		\item {\color{red}red}: rank1, {\color{blue}blue}: rank2    
		
	\end{tablenotes}
	\label{table:4}
	%	\end{threeparttable}
\end{table*}

The DTA algorithm is tested with \textit{CLEAR matrix}\cite{Sti06}. We consider few parameters: The\textit{Multi Objective Tracking Accuracy} \textbf{[MOTA]}, which counts all missed target, false positive and identity mismatches, \textit{the Multiobjective Tracking Precision}\textbf{[MOTP]} which considers the normalized distance between ground truth location and actual location. Another two parameters are, \textit{Mostly Tracked}\textbf{[MT]} and \textit{Mostly Lost}\textbf{[ML]}. Table-5 gives comparative performance of all this parameters with 5 state-of-the-art algorithms.

\begin{table*}[!ht]
	\setlength\tabcolsep{12.3pt}
	\centering
	
	\begin{tabular}{ ||c|c|c|c|c|c|c|| }
		\hline
		\textbf{Parameters} & \textbf{A}\textsuperscript{a} & \textbf{B}\textsuperscript{b} & \textbf{C}\textsuperscript{c} & \textbf{D}\textsuperscript{d} & \textbf{E}\textsuperscript{e} & \textbf{DTA} \\
		\hline\hline
		\textbf{MOTA} & 81\% & 87.6\% & \color{blue}89.5\% & \color{green}93.6\% & 77.5\% & \color{red}98.2\% \\
		\hline
		\textbf{MOTP}  & 79\% & 74.1\% & 81.1\% & \color{blue}88.6\% & \color{red} 92.8\% & \color{green}90.3\% \\
		\hline
		\textbf{MT} & - & \color{green}83.4\% & 78.2\% & 72.8\% & \color{blue}79.3\% & \color{red}84.1\% \\
		\hline
		\textbf{ML} & - & \color{red}2.3\% & \color{blue}3.5\% & 2.9\% & 3.9\% & \color{green}2.6\% \\
		\hline
		
	\end{tabular}
	\begin{tablenotes}
		\item{\textsuperscript{a}(Hsu et al.,2012),\textsuperscript{b}(Boguslaw,2014),\textsuperscript{c}(Xuan et al.,2013),\textsuperscript{d}(Zhang et al.,2014),\textsuperscript{e}(Xia et al.,2017)}
	\end{tablenotes}
	\caption{Quantitative Comparison with our proposed Dual Tracking Algorithm (DTA) and other state-of-the-art algorithm. Red, green and blue represent First, Second and Third top performance values respectively.}
	\label{table:5}
	
\end{table*}

We also perform experiment on video stream\\ \textit{Crowd\_PETS09\_S2\_L3\_Time\_14-41\_View\_01} dataset from KITTI Vision Benchmark Suite.In figure -(49) we show how successfully the proposed dual tracking algorithm(DTA) performs the tracking in comparison with other PSO algorithms.

\begin{figure*}[!ht]
	\centering
	\label{fig:49}
	\includegraphics[width=1.0\textwidth]{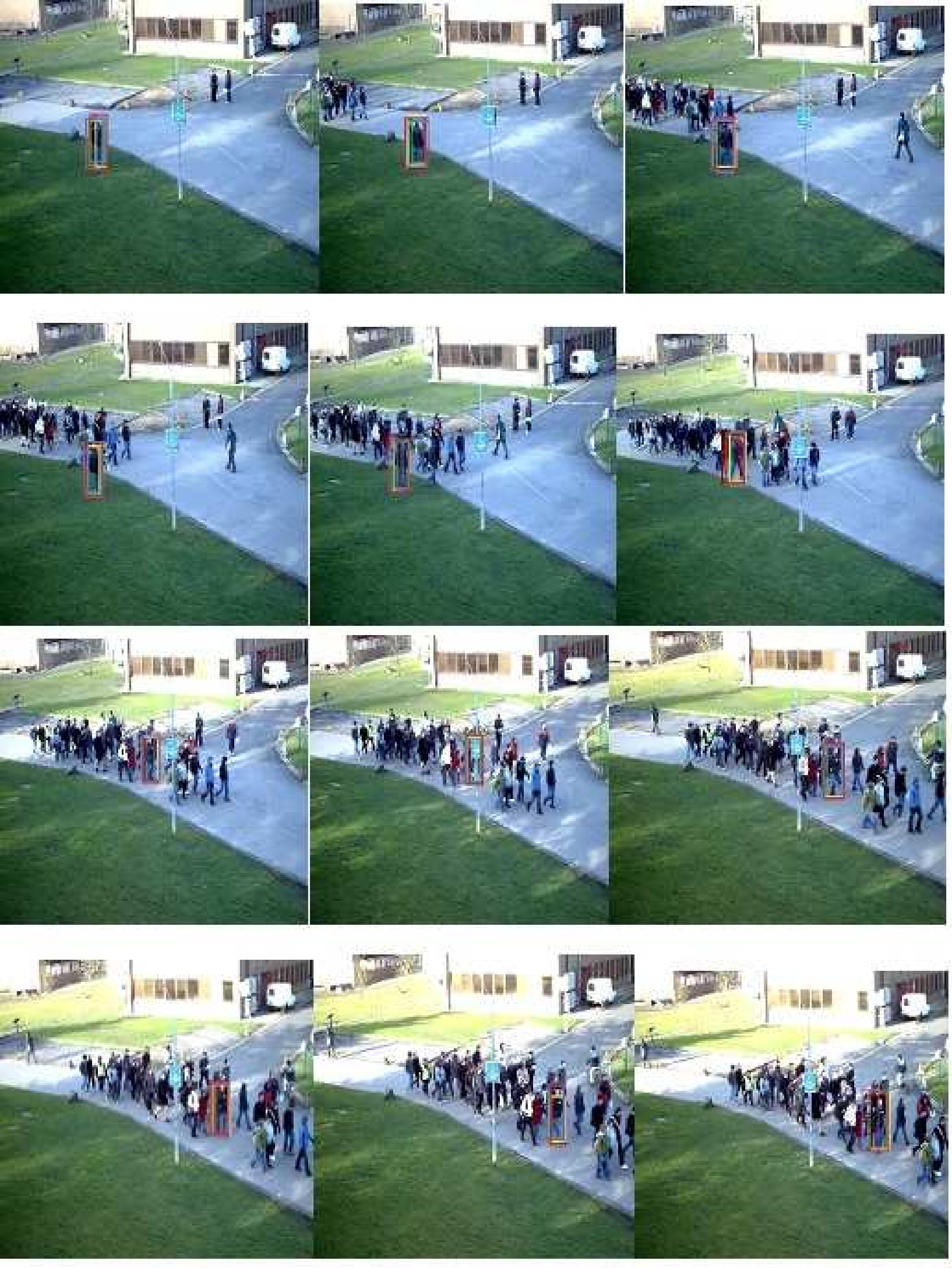}
	\caption{successful tracking images on \textit{Crowd\_PETS09\_S2\_L3\_Time\_14-41\_View\_01}. From Top to Bottom and left to right frames number 101,134,160,170,184,197,209,216,228,232,237 and 239 are tracked successfully.}
\end{figure*}

\section{Conclusion and Future Work}
\label{sec:7}
In this paper we propose a dual tracking algorithm based on optical flow and swarm intelligence. KLT tracker which tracks the dominant points of the target object is based on optical flow method whereas PSO tracker tracks the boundary information of the target object which is approximated by polygon. The proposed dual tracking algorithm (DTA) is inherently robust mainly because of two reasons; i)each tracker continuously supplement the performance of the other and thus acts as a corrective measure for each other under several disturbances during tracking as stated earlier in this paper. ii) the multiswarms annular rings(strips) where is approximated polygon of the target object is embedded captures the target object very tightly so that during tracking under several undesirable disturbances as stated earlier there is no chance for loss of tracking the target object.Hence the proposed dual tracking algorithm is robust for short video sequences and long challenging video sequence. In both the cases DTA is equally effective under static background as well as variable background.

We consider dominant point as a primary feature of the target object. It is considered as a good feature to track\cite{Shi94}. Another major advantage of choosing dominant points as good features to track is that it help constructing the approximated polygon of the target object just by joining by two consecutive dominant points. Thus from frame-2 the PSO tracker which is an important part of the dual tracking algorithm is readily supplied with approximated polygon of the target object  and a multiswarms environment is generated which provides the automatic mechanism for robustness of the dual tracking algorithm. Also the fitness function of the PSO algorithm is based on the coordinates of the dominant points. Thus dominant points of the target object have many important roles to play in dual tracking algorithm(DTA). Also it is very easy to calculate the dominant point of a new object which may arrive at any instance during tracking. If a new object appear there is no need to start the tracking from very beginning The dual tracking algorithm(DTA) will calculate the dominant point and automatically approximate the contour of the new object in the form of a polygon which will be embedded further in multiswarms annular ring(strip).Construction of the bounding box around the target object is unique and is based on the concept of PSO algorithm We test the performance of the dual tracking algorithm under several benchmark datasets and show that the performance of the proposed dual tracking algorithm(DTA) is superior then the existing algorithms as shown in section -6.The proposed dual tracking algorithm can be further improved by some finer tuning of the parameters like w, $C_1,C_2,R_1,R_2$ of the PSO tracker. The basic concepts of the multiswarms environment,with certain modification, can be further extended to object recognition and action recognition problems.

\bibliography{mybibfile}

\end{document}